\documentclass{article} % For LaTeX2e
\usepackage{iclr2017_conference,times}
\usepackage{hyperref,url}
\usepackage[pdftex]{graphicx} 
\usepackage{amsmath}
\usepackage{amssymb}
\usepackage{amsthm}
\usepackage{caption}
\usepackage{subcaption}
\usepackage{snapshot}

\iclrfinalcopy

\newcommand{\eat}[1]{}

\title{Deep Variational information bottleneck}

\author{Alexander A.~Alemi, Ian Fischer, Joshua V.~Dillon, Kevin Murphy\\
Google Research\\
\texttt{\{alemi,iansf,jvdillon,kpmurphy\}@google.com} \\
}

\usepackage{amsmath}
\usepackage{amssymb}
\usepackage{amsthm}

\newcommand{\myvec}[1]{\mathbf{#1}}
\newcommand{\myvecsym}[1]{\boldsymbol{#1}}

\newcommand{\vtheta}{\myvecsym{\theta}}

\newcommand{\vx}{\myvec{x}}

\newcommand{\vy}{\myvec{y}}

\newcommand{\vz}{\myvec{z}}
\newcommand{\mymathcal}[1]{\mathcal{#1}}
\newcommand{\calS}{\mymathcal{S}}

\newcommand{\gauss}{\mathcal{N}}

\newcommand{\softmax}{\calS}

\newcommand{\E}{\operatorname{\sf E}}
\newcommand{\sm}{\softmax}
\newcommand{\lse}{\operatorname{\sf lse}}
\newcommand{\tr}{\operatorname{tr}}
\newcommand{\diag}{\operatorname{diag}}
\newcommand{\chol}{\operatorname{chol}}
\newcommand{\subtril}{\operatorname{subtril}}

\def\tee{\textsf{\text{T}}\,}

\newcommand{\be}{\begin{equation}}
\newcommand{\ee}{\end{equation}}
\newcommand{\bea}{\begin{eqnarray}}
\newcommand{\eea}{\end{eqnarray}}
\newcommand{\beaa}{\begin{eqnarray*}}
\newcommand{\eeaa}{\end{eqnarray*}}
\newcommand{\ba}{\begin{align*}}
\newcommand{\ea}{\end{align*}}
\DeclareMathOperator{\KL}{KL}

\begin{document}

\maketitle

\begin{abstract}

We present a variational approximation to the information bottleneck of
\citet{Tishby99}.  This variational approach allows us to parameterize
the information bottleneck model using a neural network and leverage the reparameterization
trick for efficient training.  We call this method ``Deep Variational
Information Bottleneck'', or Deep VIB. We show that models trained with the
VIB objective outperform those that are trained with other forms of
regularization, in terms of generalization performance and robustness
to adversarial attack.

\end{abstract}

\section{Introduction}
\label{sec:intro}

%Information bottleneck \citep{Tishby99}.
%Reparameterization trick \citep{Kingma2014,Rezende14}.

\eat{
Empirical risk minimization entails tuning a forecaster $f(\vx; \vtheta)$ to
minimize loss over observed input/output pairs, e.g., $\min_\vtheta
\sum_i^n \ell(y_i, \hat{y}_i=f(x_i; \vtheta))$ for loss $\ell$.
}

\eat{
This paper is concerned with enhancing the interpretation of how neural
networks internally encode and decode inputs. We regard the first $n-k$ layers
of a neural network as an ``encoder;'' it maps inputs $\vx$ to some internal
representation, i.e., $\vz=f_e(\vx)$.  The $k$ final layers decode $\vz$ and
produce yet another representation; typically this representation directly
parameterizes a probability distribution over the target variable. We denote the
combined process as $f(\vx;\vtheta) = f_d(f_e(\vx;\vtheta_e);
\vtheta_d)$ and the probabilistic model is $p(y|\vx,\vtheta_e, \vtheta_d)$.
Notably the functions $f_d,f_e$ and their composition are deterministic.
}

We adopt an information theoretic view of deep networks. We regard the internal
representation of some intermediate layer as a stochastic encoding $Z$ of
the input source $X$, defined by a parametric encoder $p(\vz|\vx; \vtheta)$.\footnote{
  In this work, $X,Y,Z$ are random variables, $x,y,z$ and $\vx,\vy,\vz$ are instances
  of random variables, and $F(\cdot;\vtheta)$ and $f(\cdot;\vtheta)$ are functionals or functions
  parameterized by $\vtheta$.
}
 Our goal is to learn
an encoding that is maximally informative about our target $Y$, measured by the
mutual information between our encoding and the target $I(Z,Y; \vtheta)$, where
\be
I(Z,Y; \vtheta) = \int dx \; dy \;
p(z,y| \vtheta) \log \frac{p(z,y| \vtheta)}
{p(z| \vtheta)  p(y| \vtheta)}.\footnote{
  Note that in the present discussion, $Y$ is the ground truth label which is independent of
our parameters so $p(y|\vtheta) = p(y)$.}
\ee

Given the data processing inequality,
and the invariance of the mutual information to reparameterizations, if this
was our only objective we could always ensure a maximally informative representation
by taking the identity encoding of our data $(Z=X)$, but this is not a 
useful representation of our data. Instead we would like to find the best
 representation we can obtain subject to a constraint on its complexity.
  A natural and useful constraint to apply is on the 
mutual information between our encoding and the original data,
$I(X,Z) \leq I_c$, where $I_c$ is the information constraint. 
\eat{ Varying $I_c$ allows
us to explore a range of models with different encoding complexities, even
if they share the same architecture.  }
 This suggests the objective:
\be
	\max_{\vtheta} I(Z,Y; \vtheta) \text{ s.t. } I(X,Z; \vtheta) \leq I_c \, .
\ee
Equivalently, with the introduction of a Lagrange multiplier $\beta$,
we can maximize the objective function
\be
R_{IB}(\vtheta) = I(Z,Y; \vtheta) - \beta I(Z, X; \vtheta).
\label{eqn:IB}
\ee
Here our goal is to learn an encoding $Z$ that is maximally expressive about $Y$ 
while being maximally compressive about $X$, where $\beta \geq 0$ controls the tradeoff.\footnote{
  Note that, in our notation, large $\beta$ results in a highly compressed representation.
  In some works, the IB principle is formulated
  as the minimization of
  $I(Z,X) -\beta I(Z,Y)$,  in which case large $\beta$ corresponds to
  high mutual information between $Z$ and $Y$, and hence low compression.
}
This approach is known as the information bottleneck (IB), and was first
proposed in \citet{Tishby99}.
Intuitively, the first term in $R_{IB}$ encourages $Z$ to be predictive of $Y$; the second term
 encourages $Z$ to ``forget'' $X$.
 Essentially it forces
 $Z$ to act like a minimal sufficient statistic of $X$ for predicting $Y$.

The IB principle is appealing, since it defines what we mean by a good
representation,
in terms of the fundamental
tradeoff between having a concise representation
and one with good  predictive power
\citep{Tishby2015}.
The main drawback of the IB principle is that computing mutual
information is, in general, computationally challenging.
There are two notable exceptions:
the first is when $X$, $Y$ and $Z$ are all discrete, as in
\citet{Tishby99};
this can be used to cluster discrete data, such as words.
The second case is when $X$, $Y$ and $Z$ are all jointly Gaussian
\citep{Chechik05}.
However, these assumptions both severely constrain the class of learnable models.

In this paper, we propose to use variational inference
to construct a lower bound on
the IB objective in Equation~\ref{eqn:IB}.
We call the resulting method VIB (variational information bottleneck).
By using the reparameterization trick \citep{Kingma2014},
we can use Monte Carlo sampling to get an unbiased estimate of the gradient,
and hence we can optimize the objective
using stochastic gradient descent.
This allows us to use deep neural
networks to parameterize our distributions, and thus to handle high-dimensional,
continuous data, such as images, avoiding the previous
restrictions to the discrete or Gaussian cases.

We also show, by a series of experiments, that stochastic neural networks, fit using
our VIB method,
 are robust to overfitting,
since VIB finds a
representation $Z$ which ignores as many details of the input $X$ as
possible.
In addition,  they are more robust to adversarial inputs than deterministic models which are
fit using (penalized) maximum likelihood estimation.
Intuitively this is because
each input image gets mapped to a distribution rather than a unique $Z$,
so it is more difficult to pass small, idiosyncratic perturbations through
the latent bottleneck.

\section{Related work}
\label{sec:related}

The idea  of using information theoretic objectives for deep neural networks
was pointed out in \citet{tishby2015deep}.
However, they did not include any experimental results,
since their approach for optimizing the IB objective  relied
on the iterative Blahut Arimoto algorithm,
which is infeasible to apply to deep neural networks.

Variational inference is a natural way to approximate the problem.  Variational
bounds on mutual information have previously been explored in
\citet{agakov2004im}, though not in conjunction with the information bottleneck
objective.  \citet{Mohamed2015} also explore  variational bounds on mutual
information, and apply them to deep neural networks, but in the context of
reinforcement learning.  We recently discovered \citet{Chalk2016}, who
independently developed the same variational lower bound on the IB objective as
us.  However, they apply it to sparse coding problems, and use the kernel trick
to achieve nonlinear mappings, whereas we apply it to deep neural networks,
which are computationally more efficient.  In addition, we are able to handle
large datasets by using stochastic gradient descent, whereas they use batch
variational EM\@. Finally, concurrent with our work and most closely related
\citet{infodropout} propose and study a variational bound on the information
bottleneck objective, from the perspective of variational dropout and
demonstrate its utility in learning disentangled representations for
variational autoencoders.

In the supervised learning literature,
our work is related to the recently proposed confidence penalty (entropy regularization) method
of \citep{Pereyra16}. In this work, they fit a deterministic network
by optimizing an objective that combines the usual cross entropy
loss with an extra term which penalizes models for having low entropy
predictive distributions.
In more detail, their cost function has the form
\be
J_{CP} = \frac{1}{N} \sum_{n=1}^N \left[
  H(p(y|y_n), p(y|x_n))  - \beta H(p(y|x_n)) \right]
\label{eqn:ER}
\ee
where $H(p,q) = -\sum_y p(y) \log q(y)$ is the cross entropy,
$H(p) = H(p,p)$ is the entropy,
$p(y|y_n) = \delta_{y_n}(y)$ is a one-hot encoding of the label $y_n$,
and $N$ is the number of training examples.
(Note that setting $\beta=0$ corresponds to the usual maximum
likelihood estimate.)
In \citep{Pereyra16} they show that CP performs
better than the simpler technique of label smoothing, in which we
replace the zeros in the one-hot encoding of the labels by $\epsilon>0$,
and then renormalize so that the distribution still sums to one.
We will compare our VIB method  to both the confidence penalty method and
label smoothing
in Section~\ref{sec:mnist}.
%See also Appendix~\ref{sec:entropyreg} for further discussion of these methods.

In the unsupervised learning literature, our work is closely related to the work in
\citet{Kingma2014} on variational autoencoders.
In fact, their method is a special case of an unsupervised version of
the VIB,
but with the $\beta$ parameter fixed at 1.0,
as we explain in Appendix~\ref{sec:vae}.
The VAE objective, but with different
values of $\beta$, was also explored in \citet{earlyvisual}, but from
a different perspective.

The method of \citet{Wang16cca} proposes a latent variable generative
model of both $x$ and $y$; their variational lower bound is closely
related to ours, with the following differences.
First, we do not have a likelihood term for $x$, since we are in the
discriminative setting.
Second, they fix $\beta=1$, since they do not consider compression.

Finally, 
the variational fair autoencoder of \citet{Louizos2016}
shares with our paper the idea of ignoring  parts of the input.
However, in their approach, the user must specify which aspects of the
input (the so-called  ``sensitive'' parts) to
ignore,
whereas in our method, we can discover irrelevant parts of the input automatically.

%https://docs.google.com/document/d/1HDBFe9QRCdULSXmRRKgsuAVzBEBEhVo3gkOwXQLxDTQ/edit#

\section{Method}
\label{sec:method}

Following standard practice in the IB literature,
we assume that the joint distribution $p(X,Y,Z)$ factors as follows:
\be
p(X,Y,Z) = p(Z|X,Y) p(Y|X) p(X) = p(Z|X) p(Y|X)  p(X)
\ee
i.e., we assume $p(Z|X,Y) = p(Z|X)$, corresponding to the Markov chain
$Y \leftrightarrow X \leftrightarrow Z$.
This restriction means that our representation $Z$ cannot depend directly
on the labels $Y$.
(This opens the door to unsupervised representation learning, which we
will discuss in Appendix~\ref{sec:vae}.) Besides the structure in the joint data distribution $p(X,Y)$,
the only content at this point is our model for the stochastic encoder $p(Z|X)$, all other
distributions are fully determined by these and the Markov chain constraint.

Recall that the IB objective has the form
$I(Z,Y) - \beta I(Z,X)$. We will examine each of these expressions in turn.
Let us start with $I(Z,Y)$. Writing it out in full, this becomes
\be
I(Z,Y) = \int dy \, dz \, p(y,z) \log \frac{ p(y,z)}{p(y) p(z)}
= \int dy \, dz \, p(y, z) \log \frac{p(y|z)}{p(y)} \, .
\ee
where $p(y|z)$ is fully defined by our encoder and Markov Chain as follows:
\be
	p(y|z) = \int dx \, p(x,y|z) = \int dx\, p(y|x)p(x|z) = \int dx\, \frac{p(y|x) p(z|x) p(x)}{p(z)}.
\ee
Since this is intractable in our case,
let $q(y|z)$ be a variational approximation to $p(y|z)$.  This is our decoder, which
we will take to be another neural network with its own set of parameters.
Using the fact that the Kullback Leibler divergence is always
positive, we have
\be
\KL[ p(Y|Z), q(Y|Z) ] \geq 0
\implies \int dy \, p(y|z) \log p(y|z) \geq \int dy \, p(y|z) \log
q(y|z) \, ,
\ee
and hence
\begin{align}
  I(Z,Y)
&\geq \int dy \, dz \, p(y, z) \log \frac{ q(y|z) }{ p(y) } \\
 &= \int dy \, dz \, p(y, z) \log q(y|z) - \int dy \, p(y) \log p(y) \\
  &= \int dy \, dz \, p(y,z) \log q(y|z) + H(Y) \, .
\label{eqn:foo}
\end{align}
Notice that the entropy of our labels $H(Y)$ is independent of our optimization procedure
 and so can be ignored.

Focusing on the first term in Equation~\ref{eqn:foo},
we can rewrite  $p(y,z)$ as
$p(y,z) = \int dx \, p(x,y,z) = \int dx \, p(x) p(y|x) p(z|x)$ (leveraging
our Markov assumption),
which gives us a new lower bound on the first term of our objective:
\be
I(Z,Y) \geq \int dx \, dy \, dz \, p(x) p(y|x)   p(z|x) \log q(y|z) \, .
\ee
This only requires samples from both our joint data distribution as well as samples
from our stochastic encoder, while it requires we have access to a tractable variational
approximation in $q(y|z)$.

We now consider the term $\beta I(Z,X)$:
\be
 I(Z,X) = \int dz \, dx \, p(x,z) \log \frac{p(z|x)}{p(z)}
=  \int dz \, dx \, p(x,z) \log p(z|x)
-  \int dz \, p(z) \log p(z) \, .
\ee
In general, while it is fully defined, computing the marginal distribution of $Z$,
$p(z) = \int dx \, p(z|x) p(x)$,
might be difficult.
So let $r(z)$ be a variational approximation to this marginal.
Since
$\KL[ p(Z), r(Z) ] \geq 0 \implies \int dz \,
p(z) \log p(z) \geq \int dz \, p(z) \log r(z)$,
we have the following upper bound:
\be
 I(Z,X)
 \leq  \int dx \, dz \, p(x)  p(z|x) \log
 \frac{p(z|x)}{r(z)} \, .
\ee

Combining both of these bounds we have that
\begin{align}
  I(Z,Y) - \beta I(Z,X)
  &\geq \int dx \, dy \, dz \, p(x)  p(y|x)  p(z|x) \log q(y|z)
  \nonumber \\
  & \quad - \beta \int dx \, dz \, p(x)  p(z|x) \log \frac{ p(z|x)
  }{r(z)}
   = L \, .
  \label{eqn:bound}
\end{align}

We now discuss how to compute the lower bound $L$ in practice.
We can approximate
 $p(x,y)=p(x)p(y|x)$
using the empirical data distribution
$p(x,y) =\frac{1}{N} \sum_{n=1}^N \delta_{x_n}(x) \delta_{y_n}(y)$,
and hence we can write
\begin{equation}
  L \approx
	\frac 1 N \sum_{n=1}^N \left[ \int dz \,
           p(z|x_n) \log q(y_n|z) - \beta \, p(z|x_n) \log \frac{p(z|x_n)}{r(z)} \right].
\end{equation}

Suppose we use an encoder of the form
$p(z|x) = \gauss(z|f_e^{\mu}(x), f_e^{\Sigma}(x))$,
where $f_e$ is an MLP which outputs both the $K$-dimensional mean $\mu$ of $z$
as well as the $K \times K$ covariance matrix $\Sigma$.
Then we can use  the reparameterization trick \citep{Kingma2014}
to write
$p(z|x) dz = p(\epsilon) d\epsilon$,
where $z = f(x, \epsilon)$ is a deterministic function of $x$ and the Gaussian
random variable $\epsilon$.
\eat{
Hence  we can transform the objective to
\begin{equation}
L = 	\frac 1 N \sum_{n=1}^N \left[ \int d\epsilon \, p(\epsilon)
  \log q(y_n|z_{n,\epsilon}) - \beta \, p(z_{n,\epsilon}|x_n)
  \log \frac{p(z_{n,\epsilon}|x_n)}{r(z_{n,\epsilon})} \right].
\end{equation}
}
This formulation has the important advantage that the noise term is
independent of the parameters of the model, so it is easy to take
gradients.

Assuming our choice of $p(z|x)$ and $r(z)$ allows computation of an analytic
Kullback-Leibler divergence, we can put everything together to get the
following objective function, which we try to minimize:
\be
J_{IB} = \frac{1}{N} \sum_{n=1}^N
  \mathbb{E}_{\epsilon\sim p(\epsilon)} \left[ -\log q(y_n|f(x_n,\epsilon)) \right] +
    \beta \KL \left[ p(Z|x_n), r(Z) \right] .
\ee

\eat{
\be
J_{IB} = \frac{1}{N} \sum_{n=1}^N \left[
  \mathbb{E}_{\epsilon\sim p(\epsilon)}
  \left[ H(p(y|y_n), q(y|z = f(x_n, \epsilon)))\right] + \beta \, p(z|x_n) \log \frac{p(z|x_n)}{r(z)} \right]
\ee
where we exploit the fact that, in the case of one-hot encodings,
$H(p(y|y_n), q(y|z)) = -\int dy \, p(y|y_n) \log q(y|z) = -\log q(y_n|z)$.
}

As in \citet{Kingma2014}, this formulation allows us to directly backpropagate through a single
sample of our stochastic code and ensure that our gradient is an unbiased
estimate of the true expected gradient.\footnote{
 Even if our choice of encoding
 distribution and variational prior do not admit an analytic KL, we could
 similarly reparameterize through a sample of the divergence
 \citep{Kingma2014, blundell2015weight}.
}

\section{Experimental results}
\label{sec:results}

In this section, we present various experimental results, comparing the 
behavior of standard deterministic networks
to stochastic neural networks trained by optimizing the VIB
objective.
%For simplicity, we
%restrict attention to the well-known MNIST dataset, which consists of 60,000
%28x28 images of hand-drawn digits, from 10 classes.

\subsection{Behavior on MNIST}
\label{sec:mnist}

We start with experiments on unmodified MNIST
(i.e. no data augmentation).
In order to pick a model with some ``headroom'' to improve,
we decided to use the same architecture as
in the \citep{Pereyra16} paper,
namely an MLP with fully connected layers
of the form 784 - 1024 - 1024 - 10,
and ReLu activations.
(Since we are not exploiting spatial information,
this correpsonds to the ``permutation invariant'' version of MNIST.)
The performance of this baseline is 1.38\% error.
\citep{Pereyra16} were able to improve this
to 1.17\% using their regularization technique.
We were able to improve this to 1.13\% using our technique,
as we explain below.

In our method,  the stochastic encoder
has the form $p(z|x) = \gauss(z|f_e^{\mu}(x), f_e^{\Sigma}(x))$,
where $f_e$ is an MLP
of the form $784 - 1024 - 1024 - 2K$,
where $K$ is the size of the bottleneck.
The first $K$ outputs from $f_e$ encode $\mu$,
the remaining $K$ outputs encode $\sigma$ (after a softplus transform).

The decoder is a simple logistic
regression model of the form
$q(y|z) = \softmax(y|f_d(z))$, where
$\softmax(a) = [\exp(a_c)/\sum_{c'=1}^C \exp(a_{c'})]$ is
the softmax function,
and
$f_d(z)=Wz + b$
maps the $K$ dimensional latent code to the logits
of the  $C=10$ classes.
(In later sections, we consider more complex decoders,
but here we wanted to show the benefits of VIB in a simple setting.)

Finally, we
treat $r(z)$ as a fixed $K$-dimensional spherical Gaussian,
$r(z) = \gauss(z|0,I)$.

%While the framework allows us to use an arbitrary neural
%network for the variational classifier $q(y|z)$, as well as an arbitrary tractable variational
%approximation to the marginal $r(z)$, in these experiments we went with admitably simple choices
%to highlight the utility of the objective itself.

We compare our method to the baseline MLP.
We also consider the following deterministic limit of our model,
when $\beta=0$.
In this case, 
we obtain the following objective function:
\be
J_{IB0} = -\frac{1}{N} \sum_{n=1}^N
E_{z \sim \gauss(f_e^{\mu}(x_n),   f_e^{\Sigma}(x_n))}
\left[ \log \softmax(y_n | f_d(z)  \right]
\label{eqn:IB0}
\ee
When $\beta \rightarrow 0$, we observe the VIB optimization process tends to make
$f_e^{\Sigma}(x) \rightarrow 0$, so the network becomes nearly
deterministic.
In our experiments
we also train an explicitly deterministic model
that has the same form as the stochastic model,
except that we just use $z=f_e^{\mu}(x)$ as the hidden encoding,
and drop the Gaussian layer.

\subsubsection{Higher dimensional embedding}
\label{sec:mnist256}

To demonstrate that our VIB method can achieve competitive classification
results, we compared against a deterministic MLP trained with various
forms of regularization. We use a $K=256$ dimensional bottleneck
and a diagonal Gaussian for $p(z|x)$.
The networks were trained using
TensorFlow for 200 epochs using the Adam optimizer \citep{kingma2014adam} with a
learning rate of 0.0001. Full hyperparameter details can be found in Appendix~\ref{sec:hyperparameters}.

\begin{table}
\begin{centering}
\begin{tabular}{|r|l|}
	\hline
	Model & error \\
	\hline
	Baseline &  1.38\% \\
	Dropout &  1.34\% \\
        Dropout \citep{Pereyra16} & 1.40\% \\
	Confidence Penalty &  1.36\% \\
        Confidence Penalty  \citep{Pereyra16} &  1.17\% \\
	Label Smoothing &  1.40\% \\
        Label Smoothing \citep{Pereyra16} &  1.23\% \\
	{\bf VIB} ($\beta = 10^{-3}$) & {\bf 1.13\%}  \\
	\hline
\end{tabular}
	\caption{Test set misclassification rate on permutation-invariant MNIST using
          $K=256$. We compare our method (VIB) to an equivalent
          deterministic model using various forms of regularization.
          The discrepancy between our results for confidence penalty and label smoothing
          and the numbers reported in \citep{Pereyra16} are
          due to slightly different hyperparameters.
}
	\label{tab:mnist256}
\end{centering}
\end{table}

\begin{figure}[ht]
  \begin{center}
    \begin{tabular}{cc}
      \includegraphics[width=0.45\linewidth]{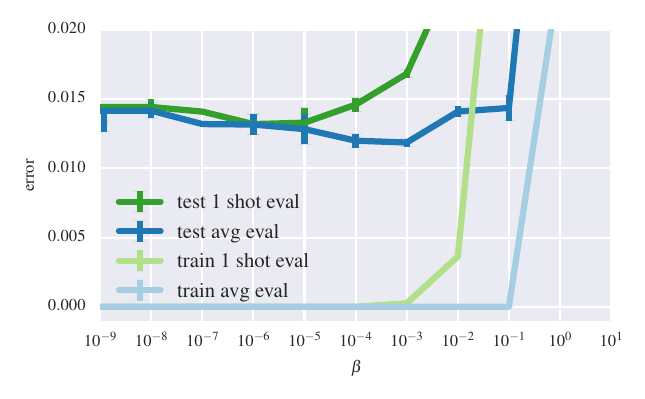}
      &
      \includegraphics[width=0.45\linewidth]{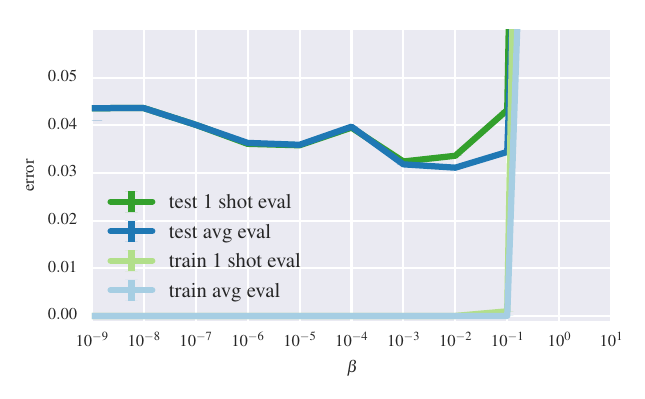} \\
      (a) & (b) \\
	\includegraphics[width=0.45\linewidth]{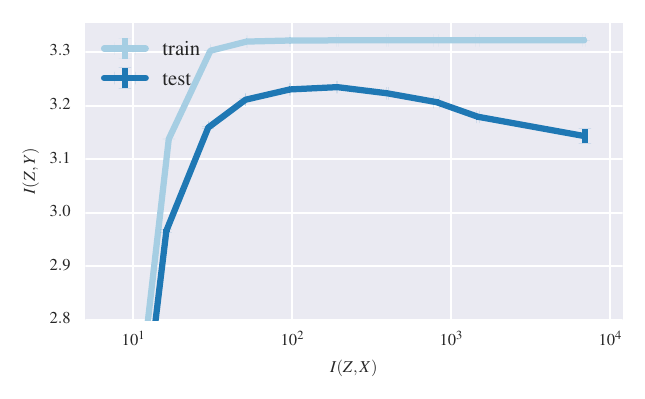}
    &
        \includegraphics[width=0.45\linewidth]{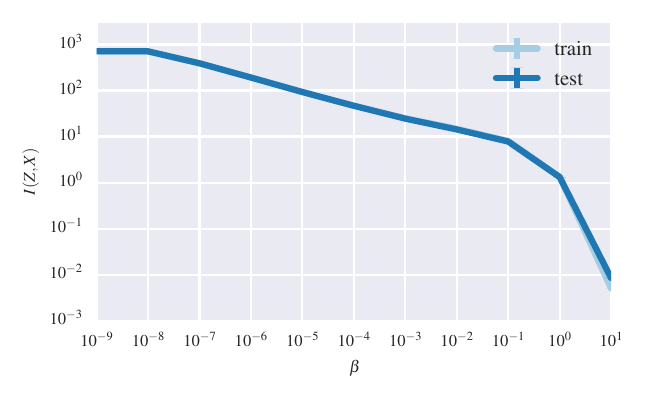}
        \\
        (c) & (d)
\end{tabular}
  \caption{Results of VIB model on MNIST.
    (a) Error rate vs $\beta$ for $K=256$ on train and test set.
    ``1 shot eval'' means a single posterior sample of $z$, ``avg eval'' means
    12 Monte Carlo samples. The spike in the error rate at
    $\beta \sim 10^{-2}$ corresponds to a model that is too highly regularized.
	  Plotted values are the average over 5 independent training runs at each $\beta$.  Error bars
	  show the standard deviation in the results.
    (b) Same as (a), but for $K=2$. Performance is much worse, since
    we pass through a very narrow bottleneck.
    (c) $I(Z,Y)$ vs $I(Z,X)$ as we vary $\beta$ for $K=256$. We see that
    increasing $I(Z,X)$ helps training set performance, but can result
    in overfitting.
    (d) $I(Z,X)$ vs $\beta$ for $K=256$.  We see that for a good value of $\beta$,
    such as $10^{-2}$, we only need to store about 10 bits of
    information about the input.
    }
	\label{fig:mnist}
\end{center}
\end{figure}

The results are shown in Table~\ref{tab:mnist256}.  We see that we can slightly
outperform other forms of regularization that have been proposed in the
literature while using the same network for each.  Of course, the performance
varies depending on $\beta$. These results are not state of the art, nor is our
main focus of our work to suggest that VIB is the best regularization method by
itself, which would require much more experimentation.  However, using the same
architecture for each experiment and comparing to VIB as the only source of
regularization suggests VIB works as a decent regularizer in and of itself.
Figure~\ref{fig:mnist}(a) plots the train and test error vs $\beta$, averaged over 5 trials (with error bars)
for the case where we use a single Monte Carlo sample of $z$ when predicting, and also
for the case where we average over 12 posterior samples (i.e., we use $p(y|x) =
\frac{1}{S} \sum_{s=1}^S q(y|z^s)$ for $z^s \sim p(z|x)$, where $S=12$). In our
own investigations, a dozen samples seemed to be sufficient to capture any
additional benefit the stochastic evaluations had to offer in this experiment\footnote{
	A dozen samples wasn't chosen for any particular reason, except the old addage that a dozen samples are sufficient,
	as mirrored in David MacKay's book~\citep{mackay}. They proved sufficient in this case.}.

We see several interesting properties in Figure~\ref{fig:mnist}(a).
First, we notice that the error rate shoots up once $\beta$ rises
above the critical value of $\beta \sim 10^{-2}$. This corresponds to
a setting where the mutual information between $X$ and $Z$ is less
than $\log_2(10)$ bits, so the model can no longer represent the
fact that there are
10 different classes.
Second, we notice that, for small values of $\beta$, the test error is higher than the training
error, which indicates that we are overfitting. This is because
the network learns to be more deterministic, forcing $\sigma \approx 0$,
thus reducing the benefits of regularization.
Third, we notice that for intermediate values of $\beta$, Monte Carlo averaging helps.
Interestingly, the region with the best performance roughly corresponds to where the added benefit from
stochastic averaging goes away, suggesting an avenue by which one could try to optimize $\beta$ using purely
statistics on the training set without a validation set.  We have not extensively studied this possibility yet.

In Figure~\ref{fig:mnist}(c), we plot the IB curve,
i.e., we plot $I(Z,Y)$ vs $I(Z,X)$ as we vary $\beta$.
As we allow more information from the input through to the bottleneck (by lowering $\beta$),
we  increase the  mutual information between our embedding and the
label on the training set,  but not necessarily on the test set, as is evident from the plot.
For more discussion about the generalization behavior of these IB curves see: \citet{generalization, clusters}.

In Figure~\ref{fig:mnist}(d) we plot the second term in our objective,
the upper bound on the mutual information between the images $X$ and our
stochastic encoding $Z$, which in our case is simply the relative entropy
between our encoding and the fixed isotropic unit Gaussian prior. Notice that
the $y$-axis is a logarithmic one.  This demonstrates that our best
results
(when $\beta$ is between $10^{-3}$ and $10^{-2}$)
occur where the mutual information between the stochastic encoding and the images
is on the order of 10 to 100 bits.
\eat{
This also means that if one were to use
our network to power a compression algorithm, our best results would require only
10 to 100 extra bits to encode each MNIST image using as a source the isotropic
unit Gaussian for the intermediate codes, while providing the best classification
results we were able to achieve here.
}

\subsubsection{Two dimensional embedding}
\label{sec:mnist2}

To better understand the behavior of our method,
we refit our model to MNIST using a $K=2$ dimensional bottleneck, but
using a full covariance Gaussian.
(The neural net predicts the mean and the  Cholesky decomposition of the
covariance matrix.)
Figure~\ref{fig:mnist}(b) shows that, not surprisingly, the classification performance is
worse (note the different scaled axes), but the overall trends are the same as in the $K=256$
dimensional case.
The IB curve (not shown) also has a similar shape to before, except
now the gap between training and testing is even larger.

Figure~\ref{fig:mnist2scatters} provides a visualization of what the network is doing.
We plot the posteriors $p(z|x)$ as a 2d Gaussian ellipse (representing
the 95\% confidence region) for 1000 images from the test set. Colors
correspond to the true class labels.
In the background of each plot is the entropy of the variational classifier
$q(y|z)$ evaluated at that point.

\newcommand{\errone}{\mbox{err}_1}
\newcommand{\errmc}{\mbox{err}_{\mbox{mc}}}

\begin{figure}[htbp]
\begin{center}
	\begin{subfigure}[b]{0.3\textwidth}
		\centering
		\includegraphics[width=1.0\textwidth]{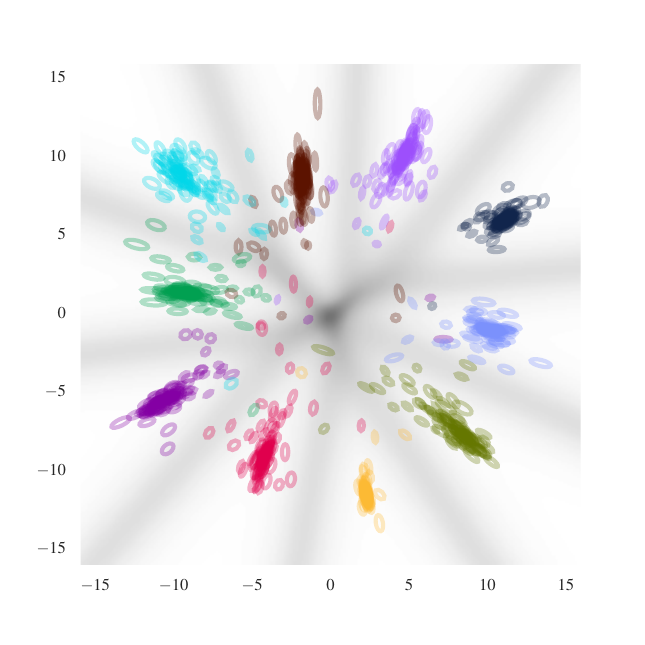}
		\caption{$\beta = 10^{-3}$, $\errmc = 3.18\%$, $\errone = 3.24\%$}
		\label{fig:mnist2scatter6}
	\end{subfigure}
	\begin{subfigure}[b]{0.3\textwidth}
		\centering
		\includegraphics[width=1.0\textwidth]{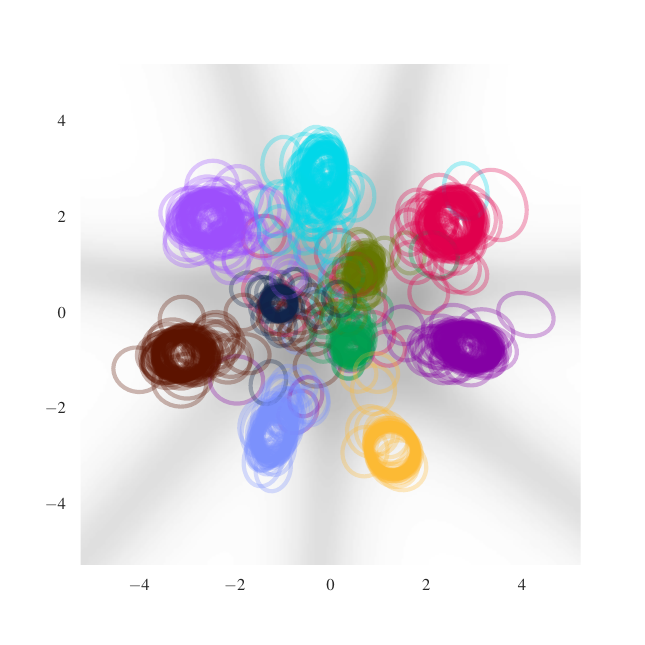}
		\caption{$\beta = 10^{-1}$, $\errmc = 3.44\%$, $\errone = 4.32\%$}
		\label{fig:mnist2scatter8}
	\end{subfigure}
	\begin{subfigure}[b]{0.3\textwidth}
		\centering
		\includegraphics[width=1.0\textwidth]{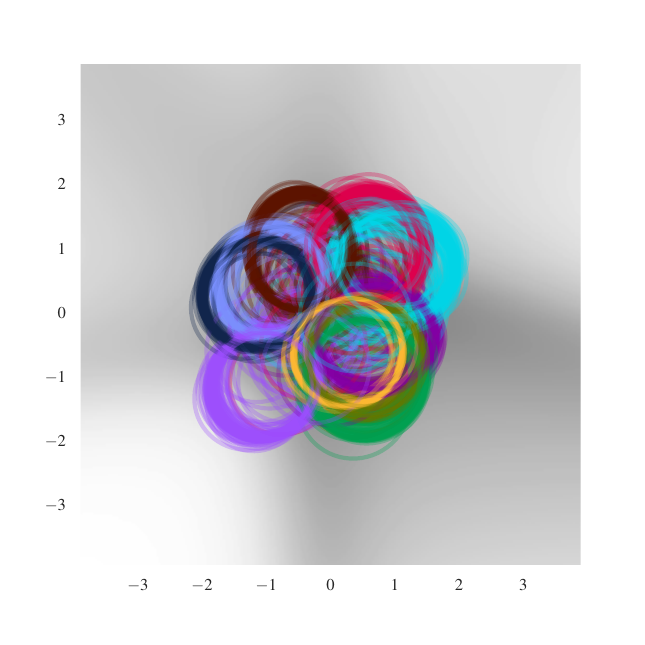}
		\caption{$\beta = 10^{0}$, $\errmc = 33.82\%$,
                  $\errone = 62.81\%$.}
		\label{fig:mnist2scatter9}
	\end{subfigure}

	\caption{Visualizing embeddings  of 1000 test images in two
          dimensions. We plot the 95\% confidence interval of the Gaussian embedding
          $p(z|x)=\gauss(\mu,\Sigma)$ as an ellipse.  The
	images are colored according to their true class label.  The
	background greyscale image denotes the entropy of the variational
	classifier evaluated at each two dimensional location. As
        $\beta$ becomes larger, we forget more about the input and
        the embeddings start to overlap to such a degree that the
        classes become indistinguishable.
        We also report the test error using a single sample, $\errone$,
        and using 12 Monte Carlo samples, $\errmc$. For ``good''
        values of $\beta$, a single sample suffices.
        }
	\label{fig:mnist2scatters}
\end{center}
\end{figure}

We see several interesting properties.
First, as $\beta$ increases (so we pass less information through), the
embedding covariances increase in relation to the distance between samples,
and the classes start to overlap.
Second, once $\beta$ passes a critical value, the encoding
``collapses'', and essentially all the class information is lost.
Third, there is a fair amount of uncertainty in the
class preditions ($q(y|z)$) in the areas between the class
embeddings.
Fourth, for intermediate values of $\beta$ (say $10^{-1}$ in
Figure~\ref{fig:mnist2scatters}(b)),
predictive performance is still good, even though there
is a lot of uncertainty about where any individual image will map to in
comparison to other images in the same class.
This means it would be difficult for an outside agent to infer which
particular instance the model is representing, a property which we
will explore more in the following sections.

\subsection{Behavior on adversarial examples}

\citet{Szegedy2013Adv} was the first work to show that deep neural
networks (and other kinds of classifiers) can be easily ``fooled''
into making mistakes by changing their inputs by imperceptibly
small amounts.
In this section, we will show
how training with the VIB objective makes models significantly more
robust to such adversarial examples.

\subsubsection{Types of Adversaries}
\label{sec:typesadv}

Since the initial work by \citet{Szegedy2013Adv} and
\citet{Goodfellow2014Adv}, many different adversaries have been
proposed.
Most attacks fall into three broad categories: optimization-based
attacks \citep{Szegedy2013Adv, Carlini2016Adv, Moosavi2015DeepFool,
  papernot2015limitations, robinson2015confusing,
  sabour2015adversarial}, which directly run an optimizer such as
L-BFGS or ADAM \citep{kingma2014adam} on image pixels to find a
minimal perturbation that changes the model's classification;
single-step gradient-based attacks \citep{Goodfellow2014Adv,
  Kurakin2016Adv, Huang2015Adv}, which choose a gradient direction of
the image pixels at some loss and then take a single step in that
direction;
and iterative gradient-based attacks \citep{Kurakin2016Adv}, which
take multiple small steps along the gradient direction of the image
pixels at some loss, recomputing the gradient direction after each
step.\footnote{
There are also other adversaries that don't fall as cleanly into those
categories, such as ``fooling images'' from \citet{Nguyen2014Fool},
which remove the human perceptual constraint, generating regular
geometric patterns or noise patterns that networks confidently
classify as natural images; and the idea of generating adversaries
by stochastic search for images near the decision boundary of multiple
networks from \citet{Baluja2015Peer}.
}

Many adversaries can be formalized as either untargeted or targeted
variants.
An untargeted adversary can be defined as $A(X, M) \rightarrow X'$,
where $A(.)$ is the adversarial function, $X$ is the input image, $X'$
is the adversarial example, and $M$ is the target model.
$A$ is considered successful if $M(X) \neq M(X')$.
Recently,
\cite{Moosavi-Dezfooli2016} showed how to create a ``universal''
adversarial perturbation $\delta$ that can be added to any image $X$
in order to make $M(X+\delta) \neq M(X)$ for a particular target model.

A targeted adversary can be defined as $A(X, M, l) \rightarrow X'$,
where $l$ is an additional target label, and $A$ is only considered
successful if $M(X') = l$.\footnote{
\citet{sabour2015adversarial} proposes a variant of the targeted
attack, $A(X_{S}, M, X_{T}, k) \rightarrow X_{S}'$, where $X_{S}$ is
the source image, $X_{T}$ is a target image, and $k$ is a target layer
in the model $M$. $A$ produces $X_{S}'$ by minimizing the difference
in activations of $M$ at layer $k$ between $X_{T}$ and $X_{S}'$.
The end result of this attack for a classification network is still
that $M(X_{S}')$ yields a target label implicitly specified by $X_{T}$
in a successful attack.
}
Targeted attacks usually require larger magnitude perturbations, since
the adversary cannot just ``nudge'' the input across the nearest
decision boundary, but instead must force it into a desired decision
region.

In this work, we focus on the Fast Gradient Sign (FGS) method proposed in
\citet{Goodfellow2014Adv} and the $L_2$ optimization method proposed in
\cite{Carlini2016Adv}.
FGS is a standard baseline attack that takes a single step in the gradient
direction to generate the adversarial example.
As originally described, FGS generates untargeted adversarial examples.
On MNIST, \citet{Goodfellow2014Adv} reported that FGS could generate adversarial
examples that fooled a maxout network approximately 90\% of the time with
$\epsilon=0.25$, where $\epsilon$ is the magnitude of the perturbation at each pixel.
The $L_2$ optimization method has been shown to generate adversarial examples
with smaller perturbations than any other method published to date, which were
capable of fooling the target network 100\% of the time.
We consider both targeted attacks and untargeted attacks for the $L_2$
optimization method.\footnote{
\citet{Carlini2016Adv} shared their code with us, which allowed us to perform
the attack with exactly the same parameters they used for their paper,
including the maximum number of iterations and maximum $C$ value (see
their paper for details).
}

\subsubsection{Adversarial Robustness}

There are multiple definitions of adversarial robustness in the literature.
The most basic, which we shall use, is accuracy on adversarially perturbed versions of the
test set, called adversarial examples.

It is also important to have a measure of the magnitude of the adversarial perturbation.
Since adversaries are defined relative to human perception, the ideal measure would explicitly correspond to how easily a human
observer would notice the perturbation.
In lieu of such a measure, it is common to compute the size of the
perturbation using $L_0$, $L_1$, $L_2$, and $L_\infty$ norms
\citep{Szegedy2013Adv, Goodfellow2014Adv, Carlini2016Adv, sabour2015adversarial}.
In particular, the $L_0$ norm measures the number of perturbed pixels,
the $L_2$ norm measures the Euclidean distance between $X$ and $X'$,
and the $L_{\infty}$ norm measures the largest single change to any pixel.

\subsubsection{Experimental Setup}

We used the same model architectures as in Section~\ref{sec:mnist},
using a $K=256$ bottleneck.
The architectures included
a deterministic (base) model trained by MLE;
a deterministic model trained with dropout (the dropout rate was
chosen on the validation set);
and a stochastic model trained with VIB for various values of $\beta$.

For the VIB models, we use 12 posterior samples of $Z$ to compute the
class label distribution $p(y|x)$.
This helps ensure that the adversaries can get a consistent gradient when constructing the perturbation, and that they can
get a consistent evaluation when checking if the perturbation was
successful (i.e., it reduces the chance that the adversary ``gets
lucky'' in its perturbation due to an untypical sample).
We also ran the VIB models in ``mean mode'', where the $\sigma$s are forced to be 0.
This had no noticeable impact on the results, so all reported results are for stochastic evaluation with 12 samples.

\subsubsection{MNIST Results and Discussion}
\label{sec:mnistadv}

\begin{figure*}[p]
  \begin{center}
    \hspace{-1cm}
    \begin{subfigure}[t]{0.95\textwidth}
      \centering
      \setlength\tabcolsep{1pt}

        \begin{tabular}{cccccccccc}
            $Orig.$ & $Det.$ & $Dropout$ & $\beta=0$ & $\beta=10^{-10}$  & $\beta=10^{-8}$ &
            $\beta=10^{-6}$ & $\beta=10^{-4}$ & $\beta=10^{-3}$ & $\beta=10^{-2}$ \\
        \includegraphics[width=1.2cm,keepaspectratio,trim=1 1 1 1,clip]{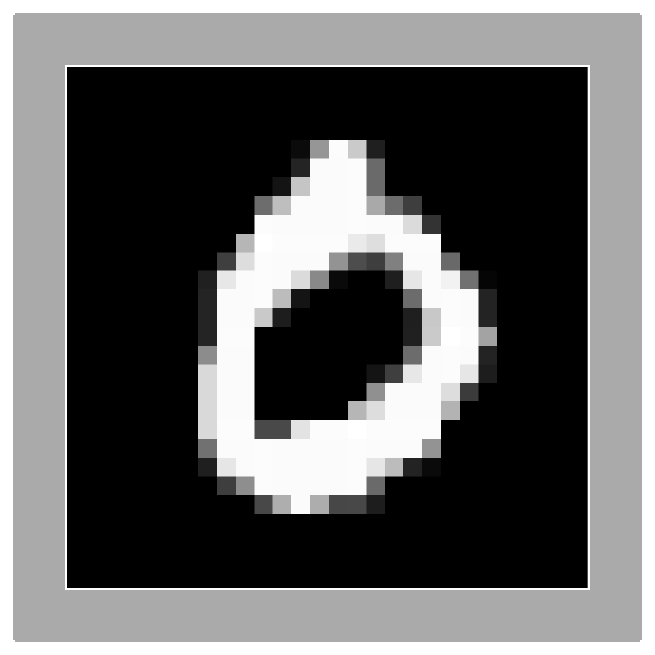} &
        \includegraphics[width=1.2cm,keepaspectratio,trim=1 1 1 1,clip]{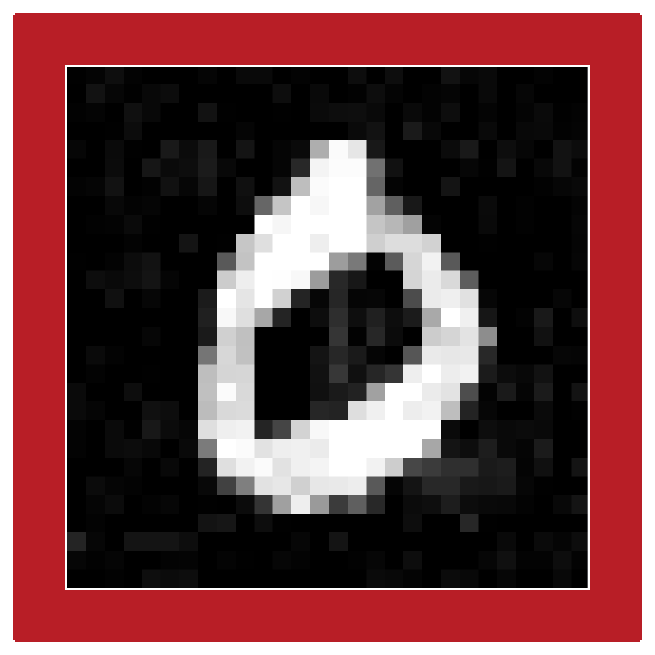} &
        \includegraphics[width=1.2cm,keepaspectratio,trim=1 1 1 1,clip]{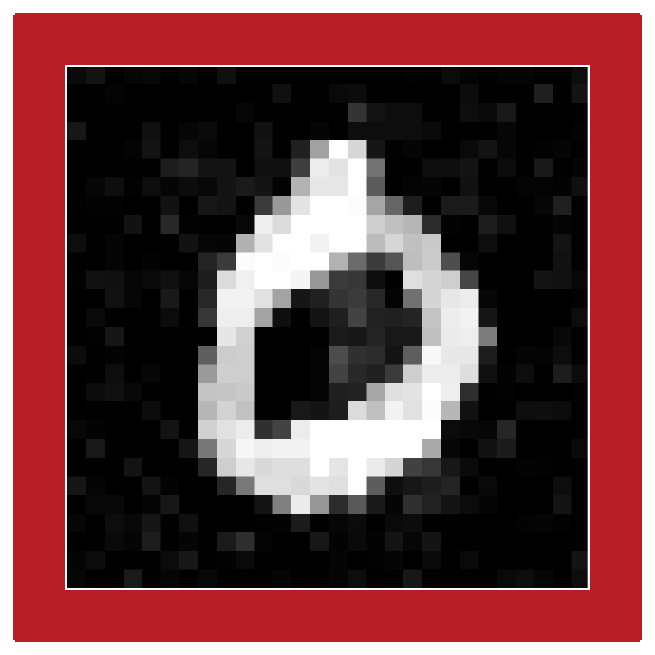} &
        \includegraphics[width=1.2cm,keepaspectratio,trim=1 1 1 1,clip]{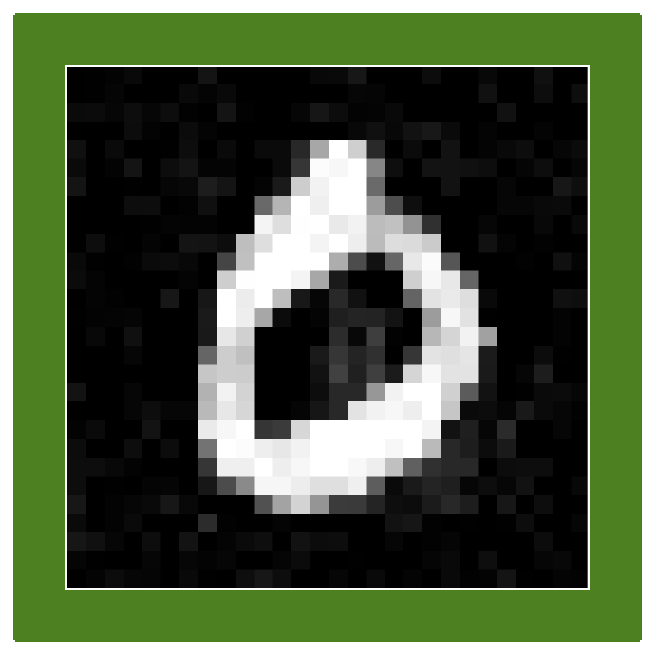} &
        \includegraphics[width=1.2cm,keepaspectratio,trim=1 1 1 1,clip]{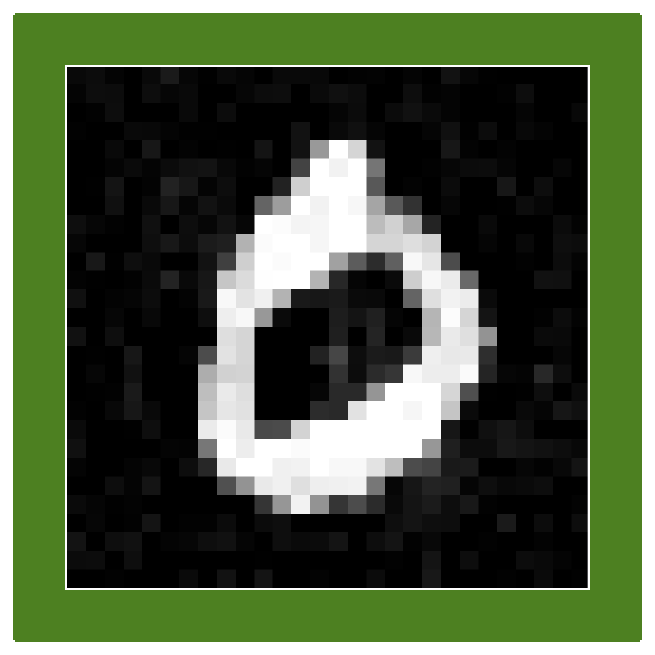} &
        \includegraphics[width=1.2cm,keepaspectratio,trim=1 1 1 1,clip]{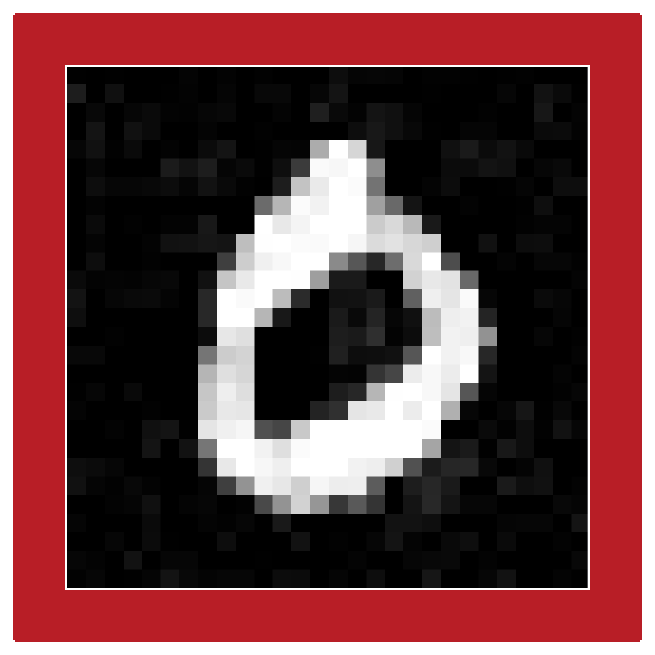} &
        \includegraphics[width=1.2cm,keepaspectratio,trim=1 1 1 1,clip]{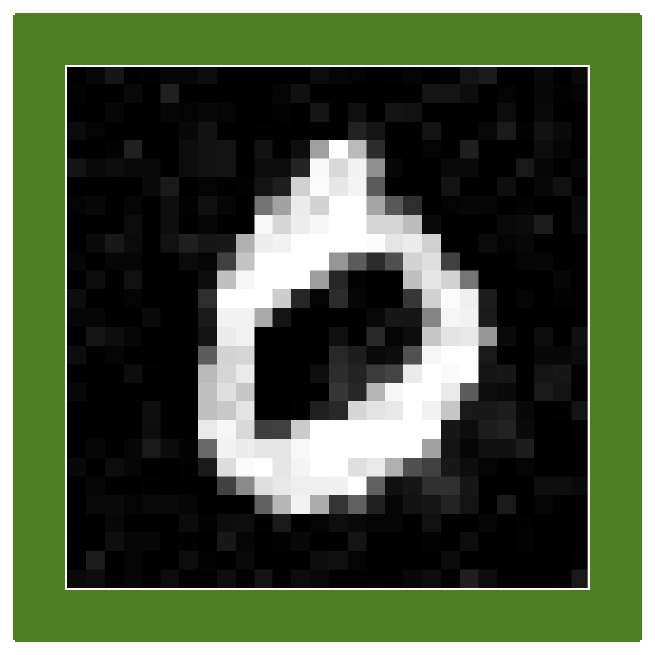} &
        \includegraphics[width=1.2cm,keepaspectratio,trim=1 1 1 1,clip]{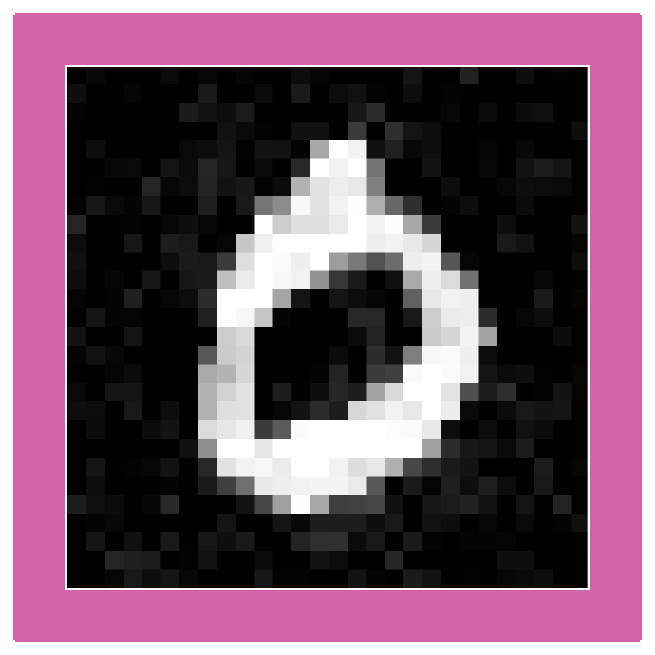} &
        \includegraphics[width=1.2cm,keepaspectratio,trim=1 1 1 1,clip]{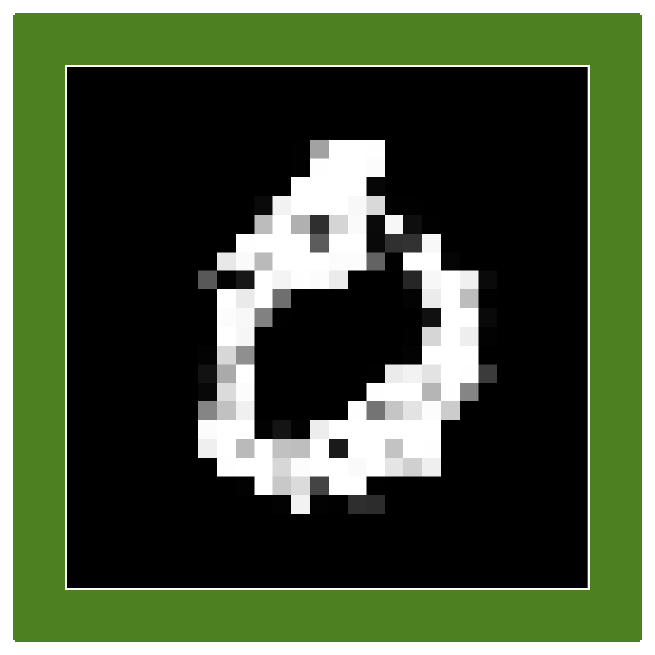} &
        \includegraphics[width=1.2cm,keepaspectratio,trim=1 1 1 1,clip]{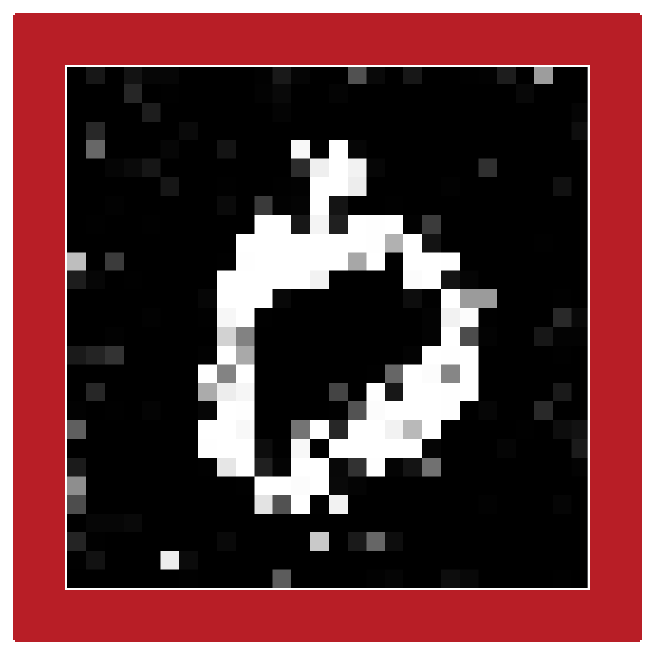} \\

        \includegraphics[width=1.2cm,keepaspectratio,trim=1 1 1 1,clip]{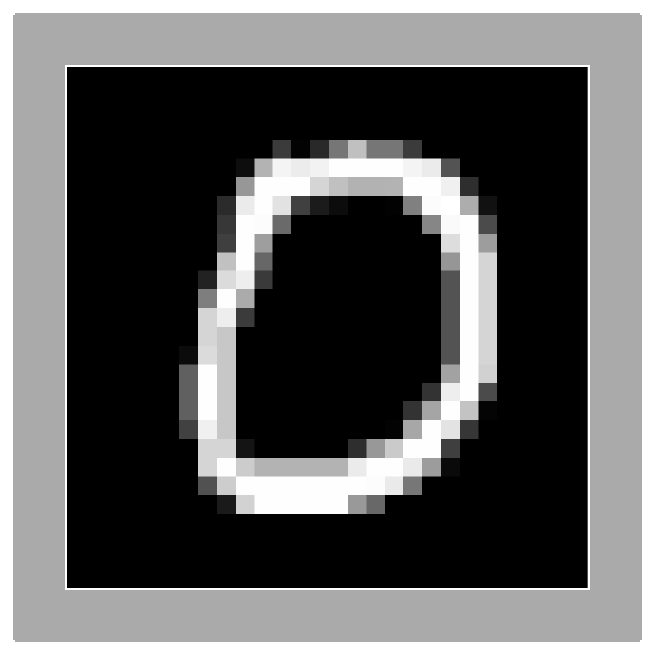} &
        \includegraphics[width=1.2cm,keepaspectratio,trim=1 1 1 1,clip]{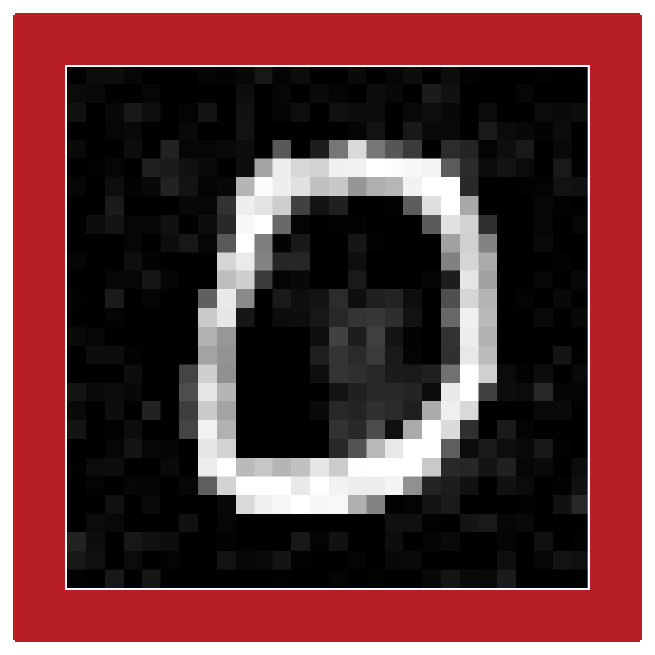} &
        \includegraphics[width=1.2cm,keepaspectratio,trim=1 1 1 1,clip]{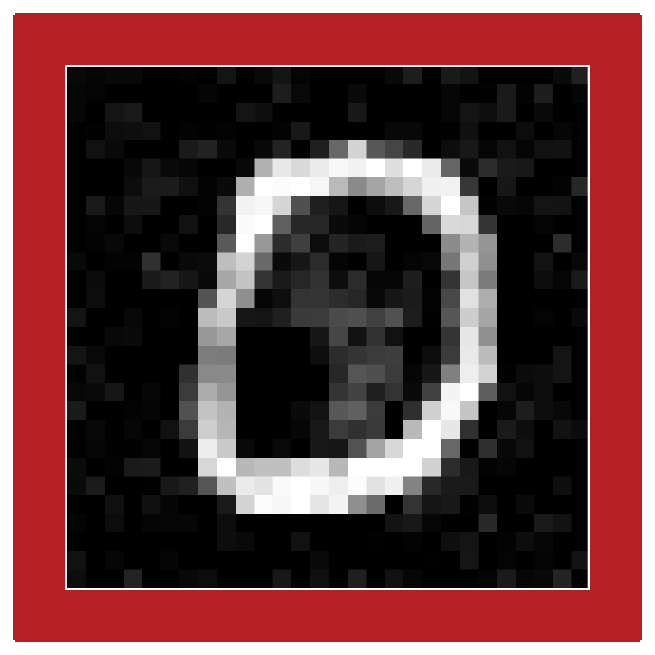} &
        \includegraphics[width=1.2cm,keepaspectratio,trim=1 1 1 1,clip]{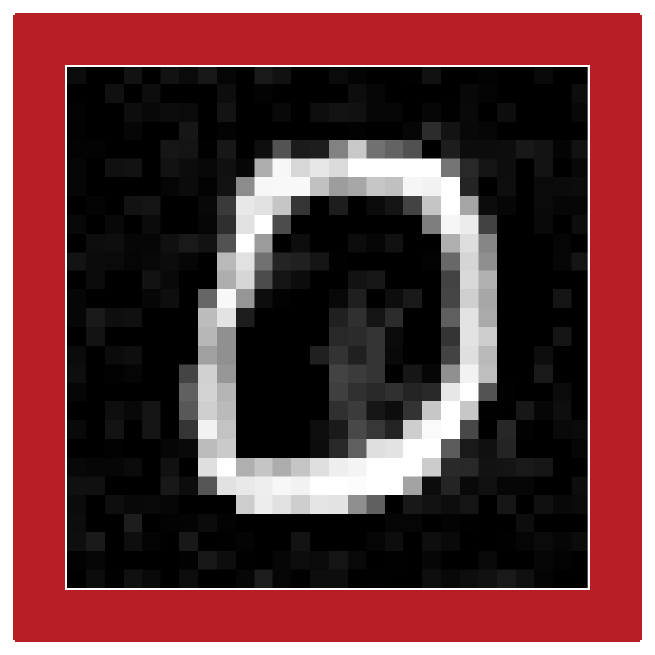} &
        \includegraphics[width=1.2cm,keepaspectratio,trim=1 1 1 1,clip]{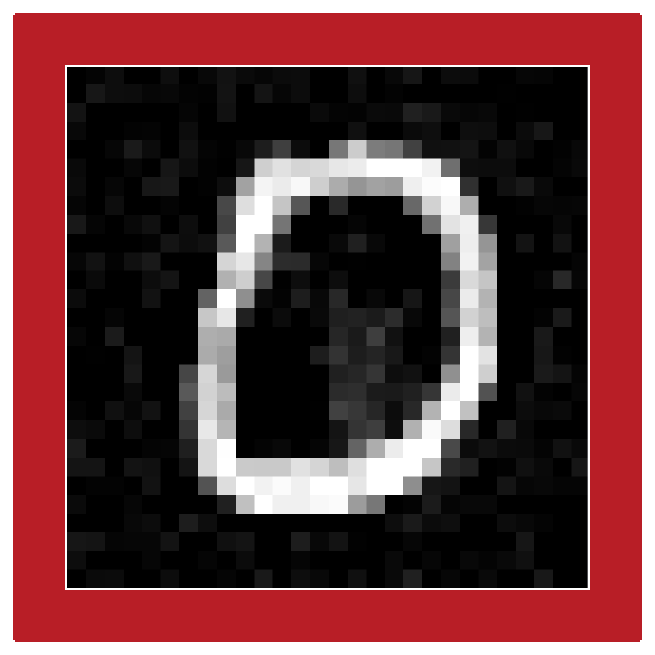} &
        \includegraphics[width=1.2cm,keepaspectratio,trim=1 1 1 1,clip]{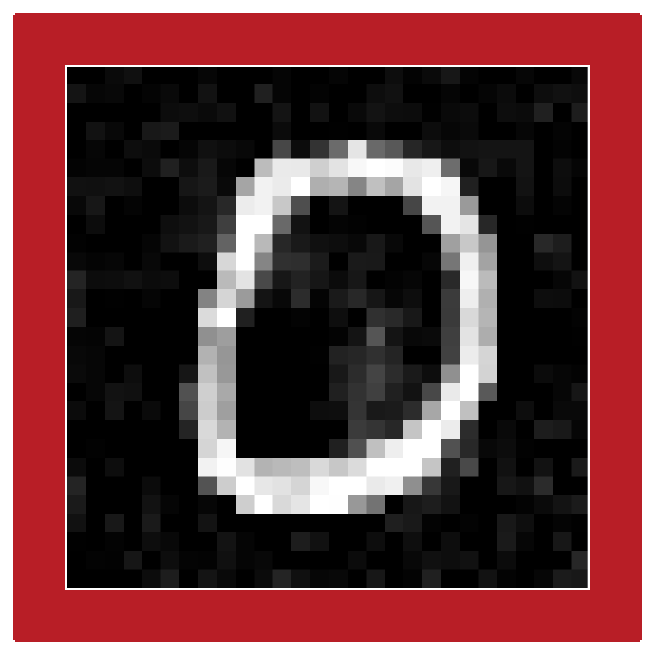} &
        \includegraphics[width=1.2cm,keepaspectratio,trim=1 1 1 1,clip]{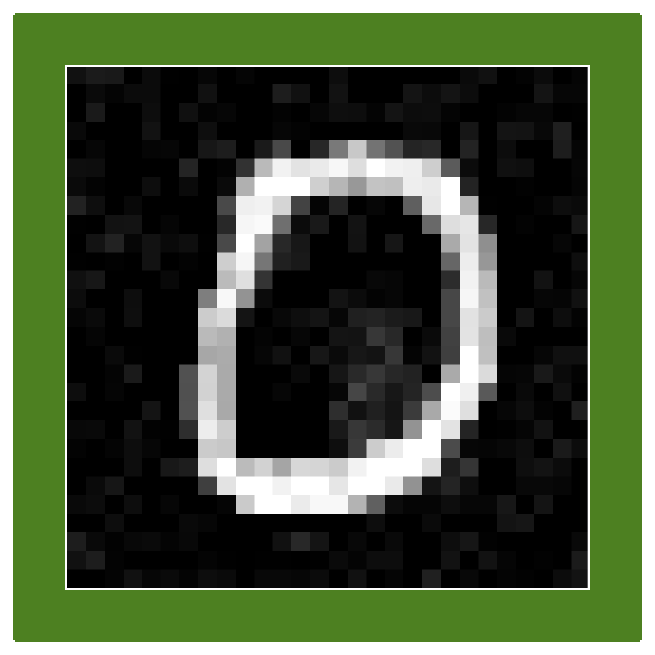} &
        \includegraphics[width=1.2cm,keepaspectratio,trim=1 1 1 1,clip]{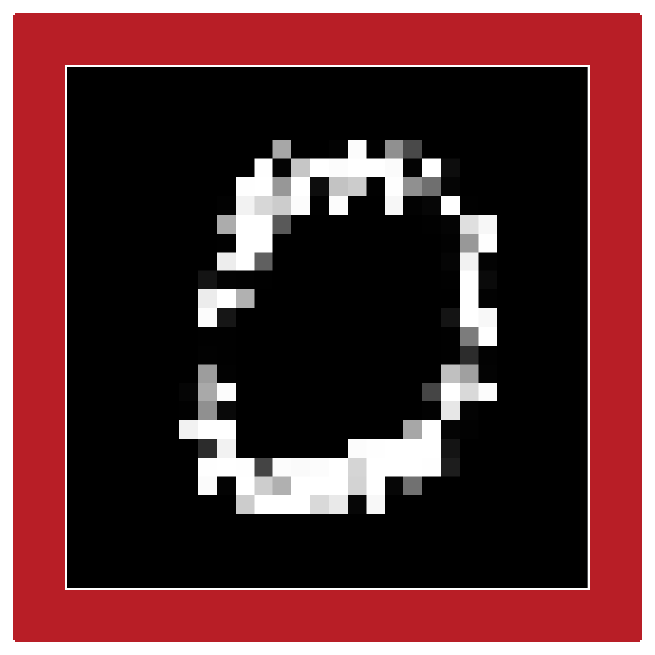} &
        \includegraphics[width=1.2cm,keepaspectratio,trim=1 1 1 1,clip]{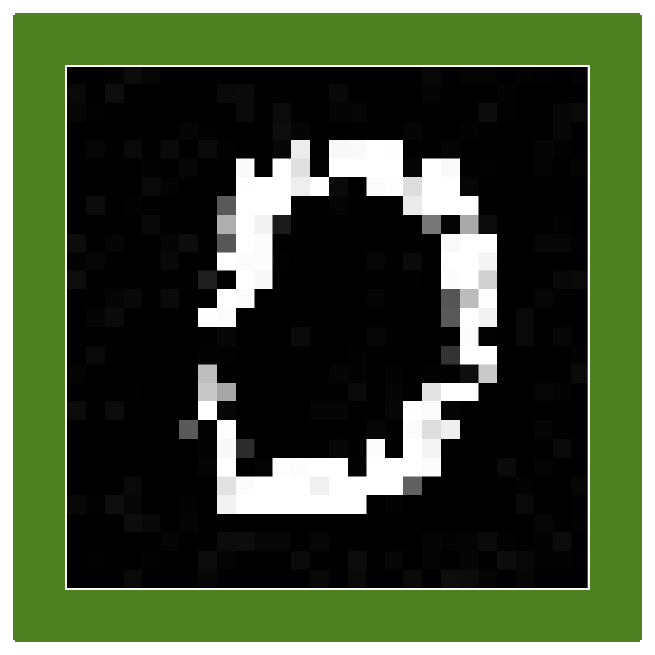} &
        \includegraphics[width=1.2cm,keepaspectratio,trim=1 1 1 1,clip]{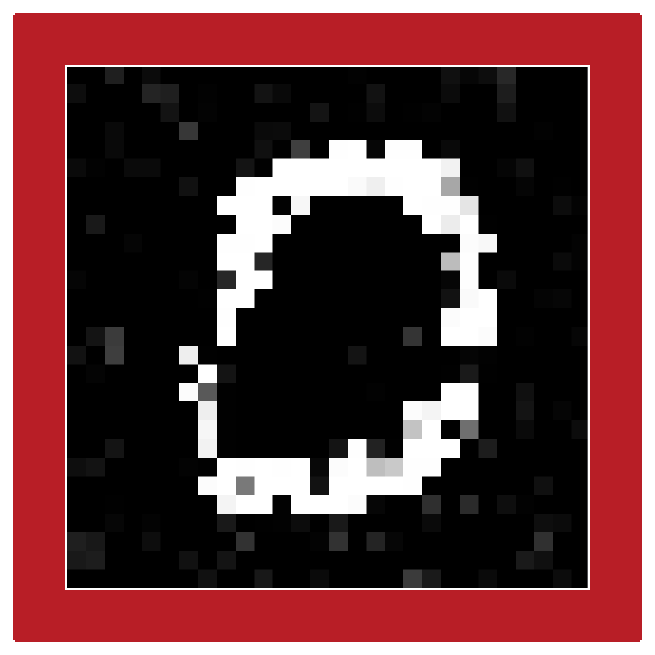} \\

        \includegraphics[width=1.2cm,keepaspectratio,trim=1 1 1 1,clip]{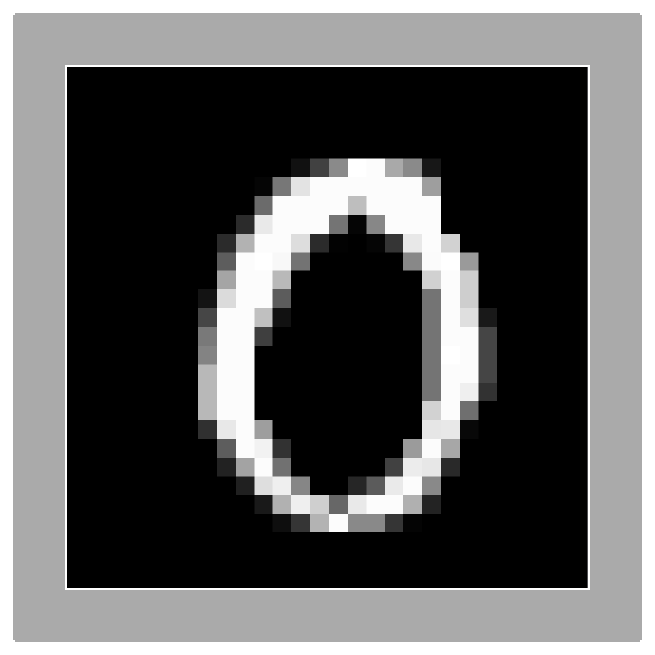} &
        \includegraphics[width=1.2cm,keepaspectratio,trim=1 1 1 1,clip]{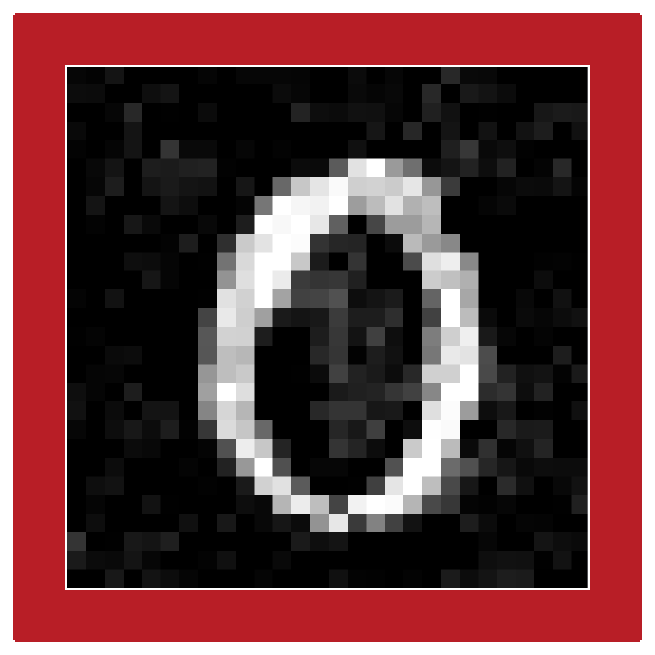} &
        \includegraphics[width=1.2cm,keepaspectratio,trim=1 1 1 1,clip]{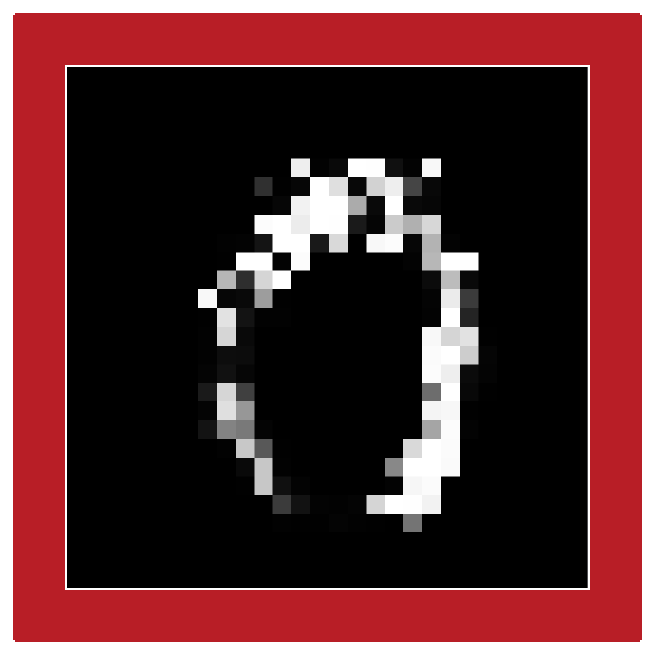} &
        \includegraphics[width=1.2cm,keepaspectratio,trim=1 1 1 1,clip]{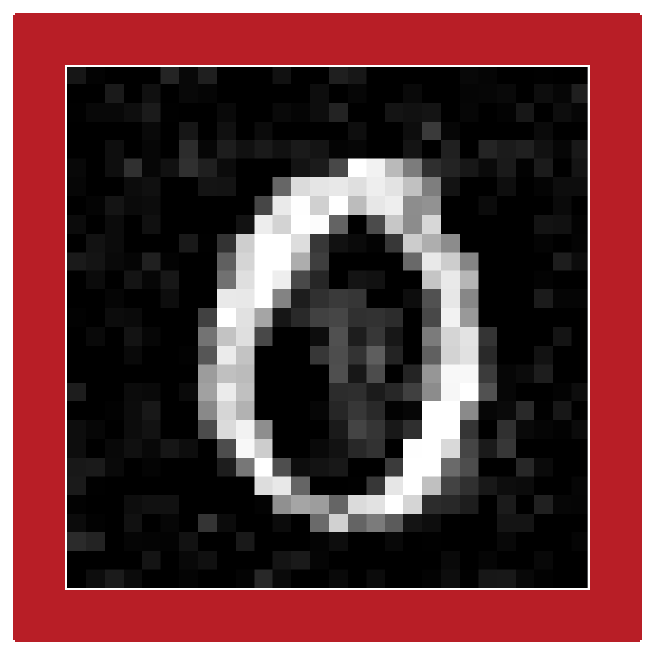} &
        \includegraphics[width=1.2cm,keepaspectratio,trim=1 1 1 1,clip]{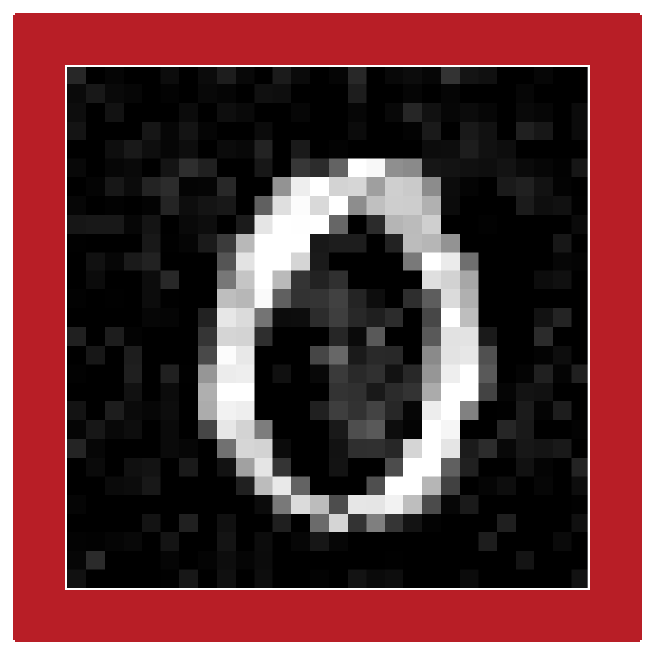} &
        \includegraphics[width=1.2cm,keepaspectratio,trim=1 1 1 1,clip]{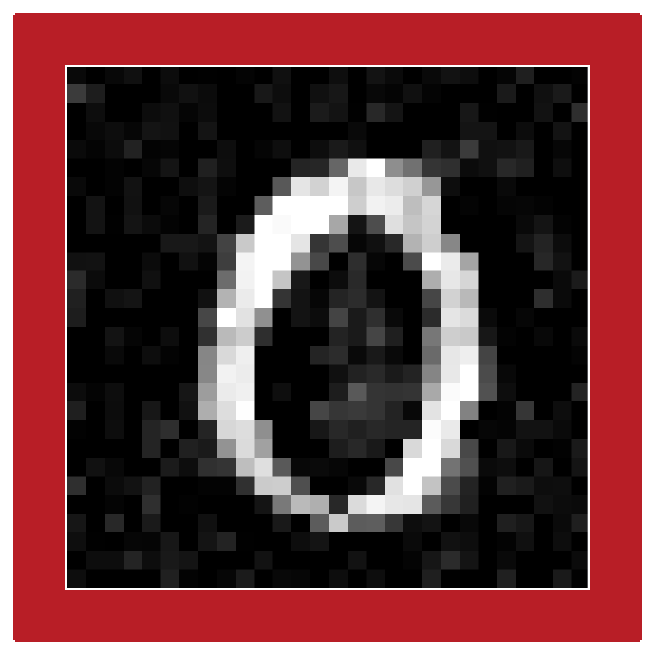} &
        \includegraphics[width=1.2cm,keepaspectratio,trim=1 1 1 1,clip]{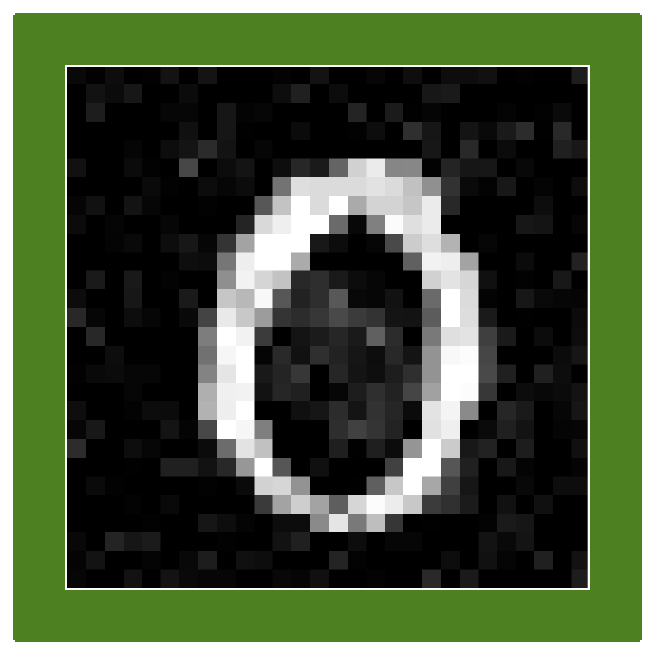} &
        \includegraphics[width=1.2cm,keepaspectratio,trim=1 1 1 1,clip]{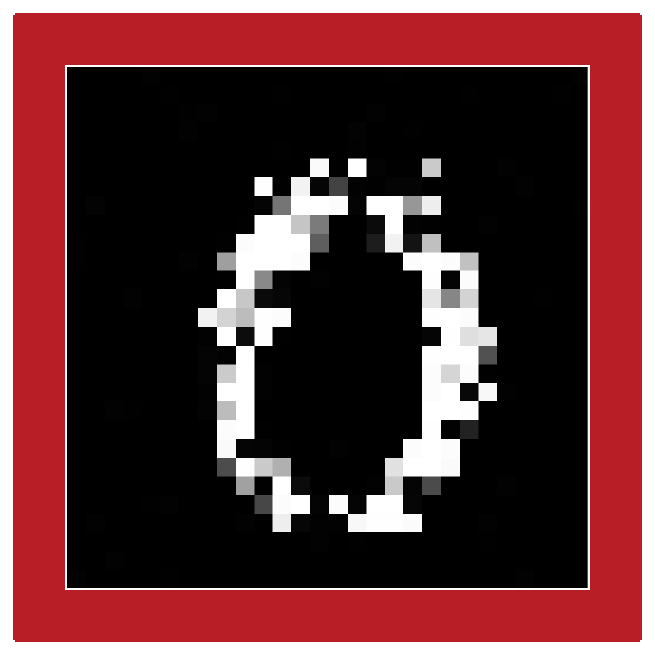} &
        \includegraphics[width=1.2cm,keepaspectratio,trim=1 1 1 1,clip]{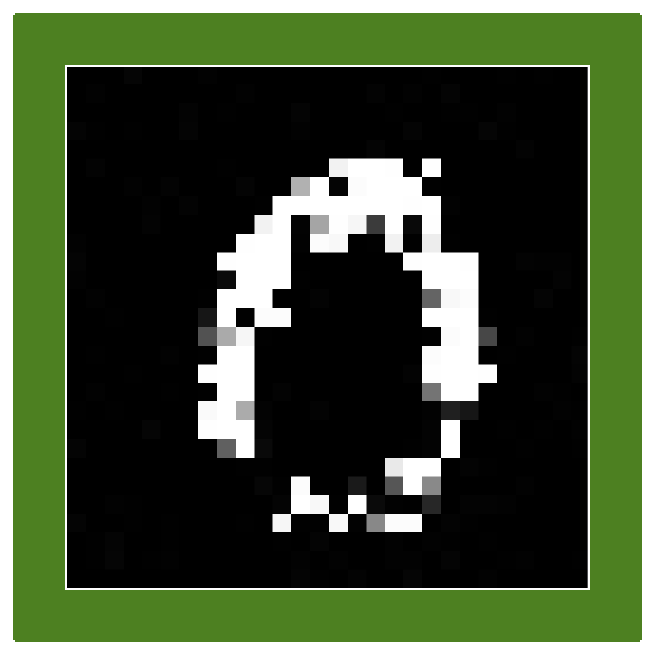} &
        \includegraphics[width=1.2cm,keepaspectratio,trim=1 1 1 1,clip]{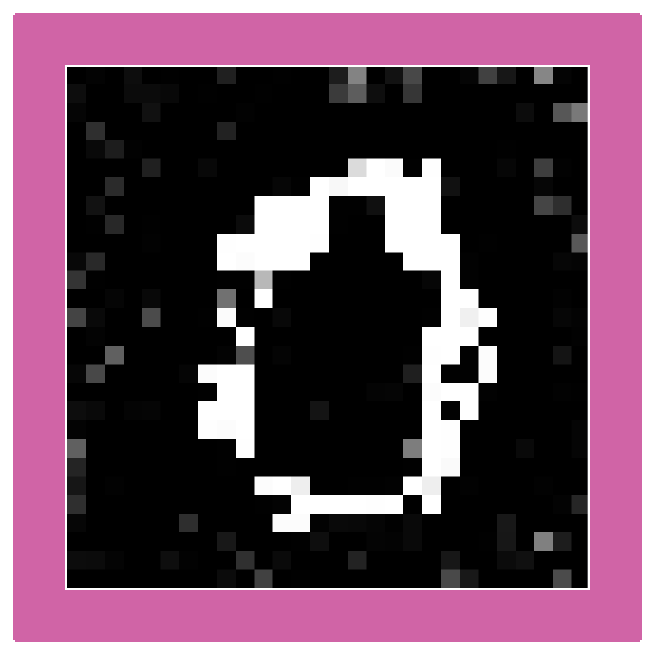} \\
        \end{tabular}
        \end{subfigure}
    \caption{The adversary is trying to force each 0 to be classified as a 1.
      Successful attacks have a red background.
      Unsuccessful attacks have a green background.
      In the case that the label is changed to an incorrect label different from the target label
      (i.e., the classifier outputs something other than 0 or 1), the background is purple.
      The first column is the original image.
      The second column is adversarial examples targeting our deterministic baseline model.
      The third column is adversarial examples targeting our dropout model.
      The remaining columns are adversarial examples targeting our VIB models for different $\beta$.
    }
    \label{fig:adv_carlini_zeros}
\end{center}
\end{figure*}

\begin{figure*}[htbp]
% \begin{center}
  % \centering
  % \hspace{-4cm}
\centerline{
  \begin{tabular}{cc}
    \includegraphics[height=6cm,keepaspectratio]{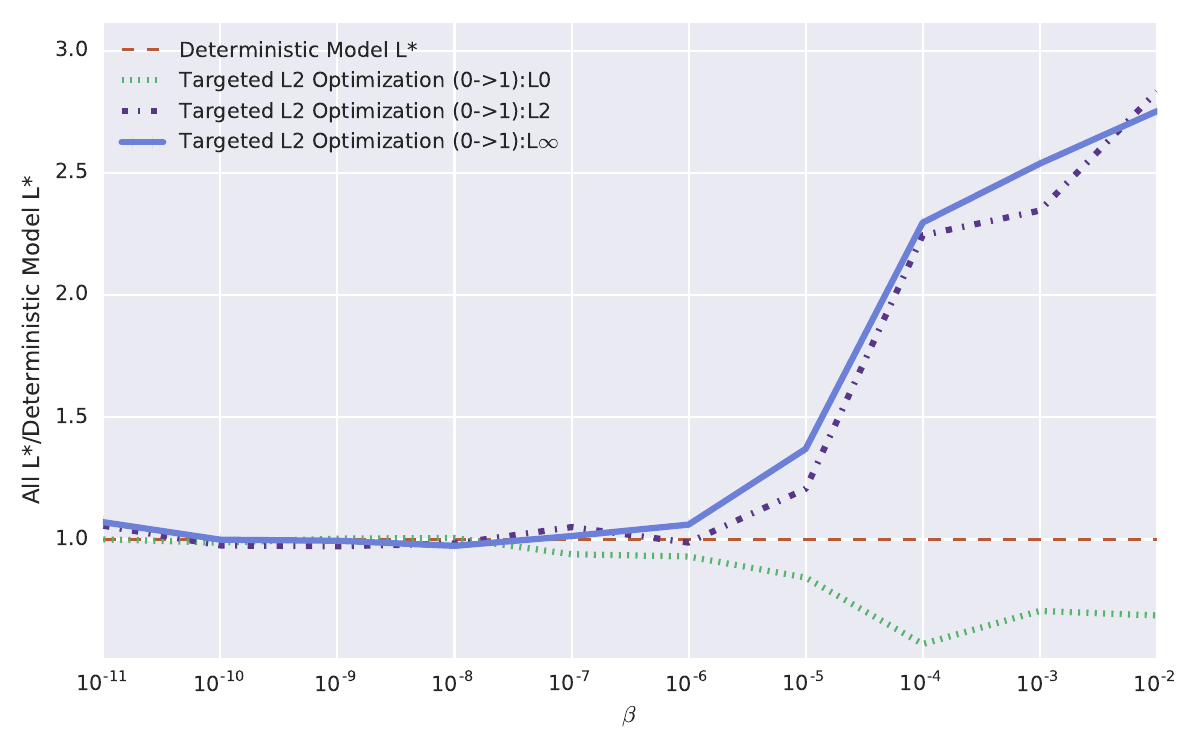}
    &
    \includegraphics[height=6cm,keepaspectratio]{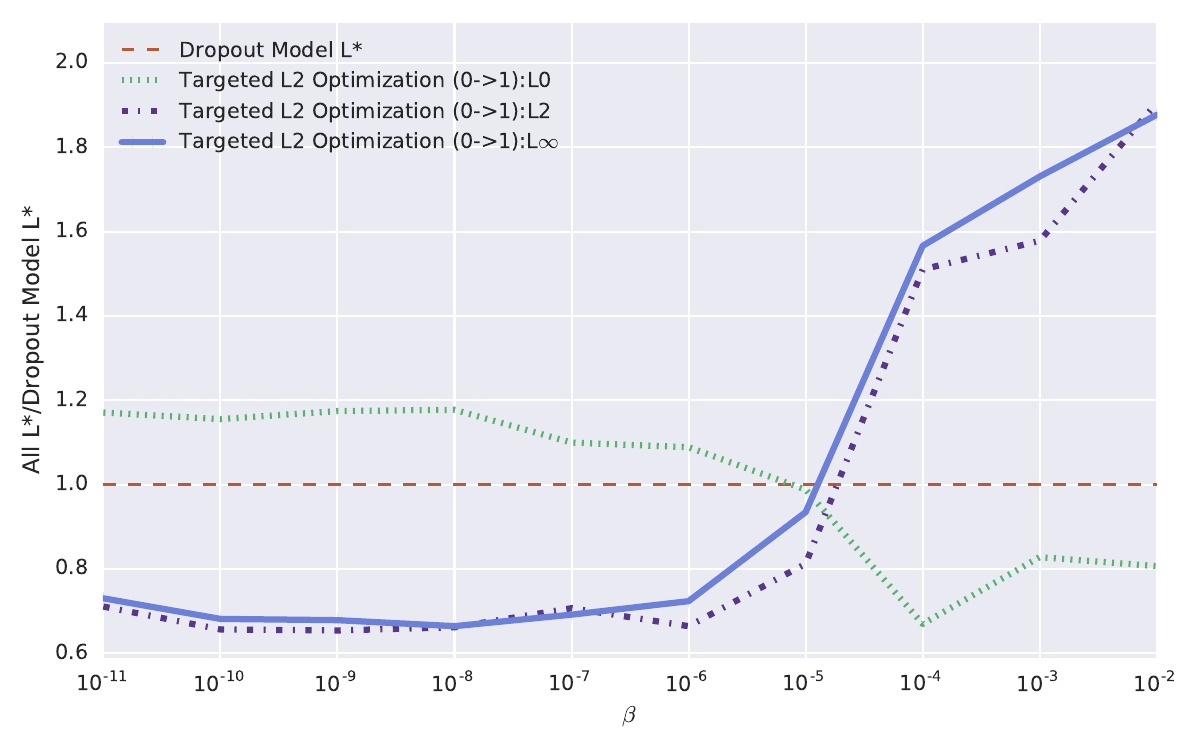}
    \\
    (a) & (b) \\
  \end{tabular}
}
  \caption{
    (a) Relative magnitude of the adversarial perturbation, measured using $L_0$, $L_2$, and $L_{\infty}$ norms,
    for the images in Figure~\ref{fig:adv_carlini_zeros} as a function of $\beta$.
    (We normalize all values by the corresponding norm of the perturbation against the base model.)
    As $\beta$ increases, $L_0$ decreases, but both $L_2$ and $L_\infty$ increase,
    indicating that the adversary is being forced to put larger modifications
    into fewer pixels while searching for an adversarial perturbation.
    (b) Same as (a), but with the dropout model as the baseline. Dropout is more robust to the adversarial perturbations
    than the base deterministic model, but still performs much worse than the VIB model as $\beta$ increases.
}
  \label{fig:adv_carlini_l_star}
% \end{center}
\end{figure*}

We selected the first 10 zeros in the MNIST test set, and use the $L_2$
optimization adversary
of \citet{Carlini2016Adv} to try to perturb those zeros into ones.\footnote{
We chose this pair of labels since intuitively zeros and ones are
the  digits that are least similar
in terms of human perception, so if the adversary can change a zero into a
one without much human-noticeable perturbation, it is unlikely that the model
has learned a representation similar to what humans learn.
}
Some sample results are shown in Figure \ref{fig:adv_carlini_zeros}.
We see that the deterministic models are easily fooled by making small
perturbations, but for the VIB models with reasonably large $\beta$,
the adversary often fails to find an attack (indicated by the green borders)
within the permitted number of iterations.
Furthermore, when an attack is succesful,
it needs to be much larger for the VIB models.
To quantify this, Figure~\ref{fig:adv_carlini_l_star} plots the magnitude of
the perturbation (relative to that of the deterministic and dropout models) needed
for a successful attack as a function of $\beta$.
As $\beta$ increases, the $L_0$ norm of the perturbation decreases, but both
$L_2$ and $L_\infty$ norms increase, indicating that the adversary is being
forced to put larger modifications into fewer pixels while searching for an
adversarial perturbation.

\begin{figure*}[htbp]
% \begin{center}
  % \centering
  % \hspace{-4cm}
\centerline{
  \begin{tabular}{cc}
    \includegraphics[height=6cm,keepaspectratio]{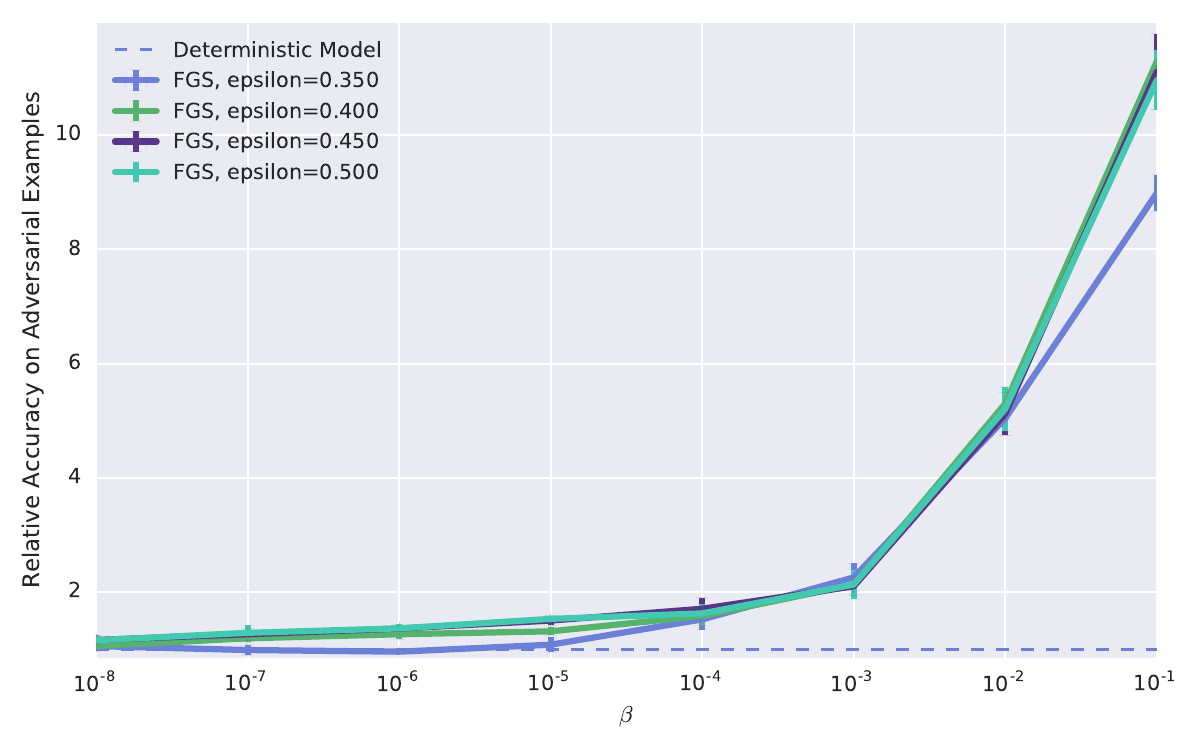}
    &
    \includegraphics[height=6cm,keepaspectratio]{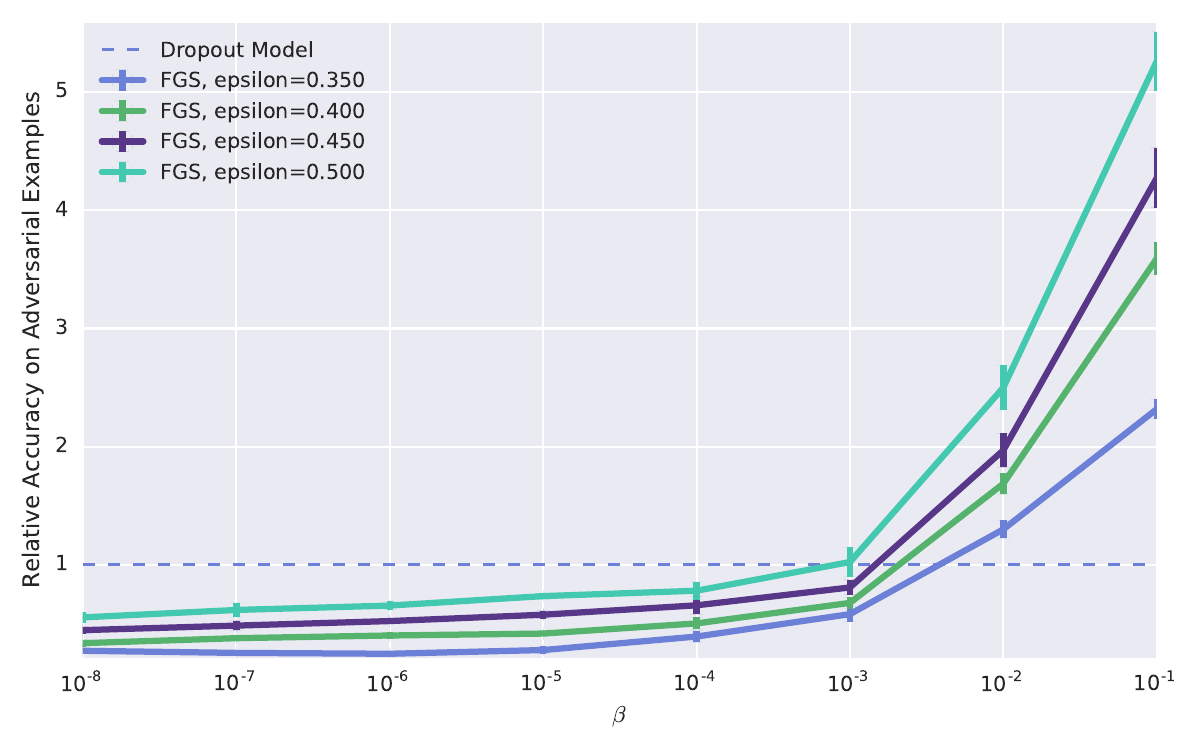}
    \\
    (a) & (b) \\
  \end{tabular}
}
  \caption{
    Classification accuracy of VIB classifiers, divided by accuracy of baseline classifiers, on FGS-generated adversarial
    examples as a function of $\beta$. Higher is better, and the baseline is always at $1.0$.
    For the FGS adversarial examples, when $\beta=0$ (not shown), the VIB model's performance is almost identical to when
    $\beta=10^{-8}$.
    (a) FGS accuracy normalized by the base deterministic model performance.
    The base deterministic model's accuracy on the adversarial examples ranges from about 1\% when $\epsilon=0.5$ to about
    5\% when $\epsilon=0.35$.
    (b) Same as (a), but with the dropout model as the baseline.
    The dropout model is more robust than the base model, but less robust than VIB, particularly for stronger adversaries
    (i.e., larger values of $\epsilon$).
    The dropout model's accuracy on the adversarial examples ranges from about 5\% when $\epsilon=0.5$ to about
    16\% when $\epsilon=0.35$.
    As in the other results, relative performance is more dramatic as $\beta$ increases, which seems to indicate that the
    VIB models are learning to ignore more of the perturbations caused by the FGS method, even though they were not
    trained on any adversarial examples.
}
  \label{fig:adv_fgs_acc}
% \end{center}
\end{figure*}

Figure~\ref{fig:adv_fgs_acc} plots the accuracy on FGS
adversarial examples of the first 1000 images from the MNIST
test set as a function of $\beta$.
Each point in the plot corresponds to 3 separate executions of three different models
trained with the same value of $\beta$.
All models tested achieve over 98.4\% accuracy on the unperturbed MNIST test set, so
there is no appreciable measurement distortion due to underlying model accuracy.

\begin{figure*}[htbp]
\centerline{
    \includegraphics[height=6cm,keepaspectratio]{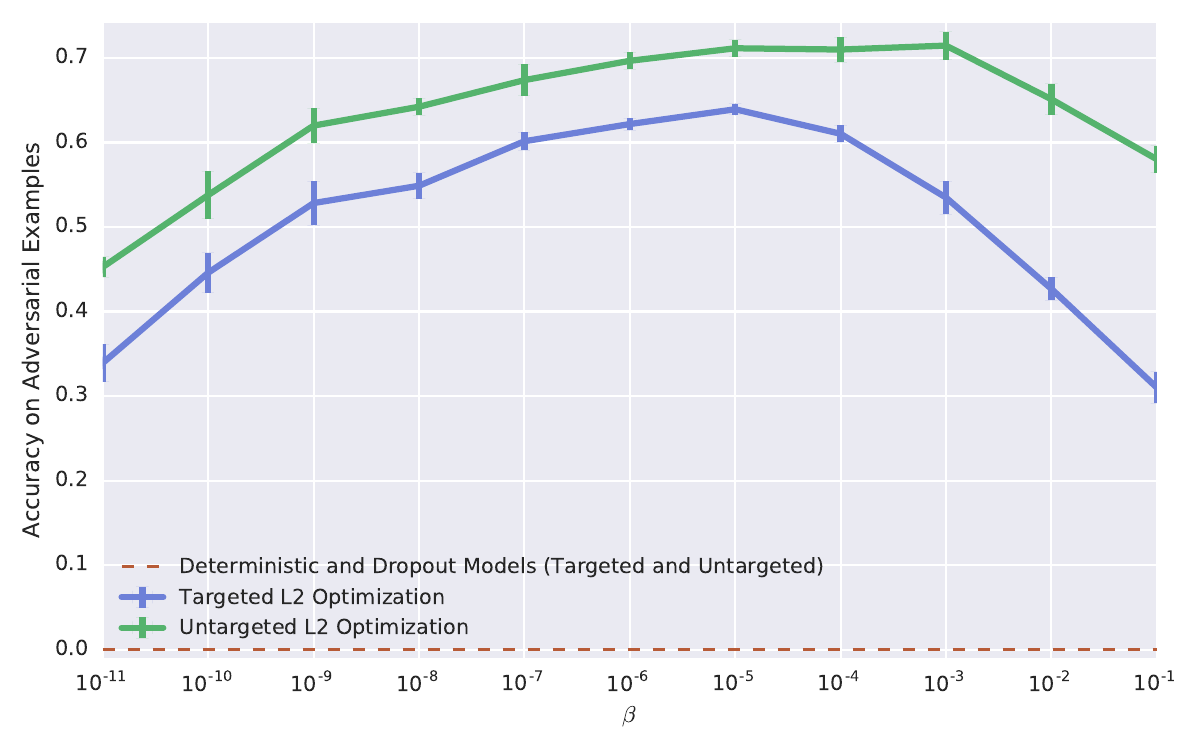}
}
  \caption{
    Classification accuracy (from 0 to 1) on $L_2$ adversarial examples (of all classes) as a function of $\beta$.
    The blue line is for targeted attacks, and the green line is for untargeted attacks (which are easier to resist).
    In this case, $\beta=10^{-11}$ has performance indistinguishable from $\beta=0$.
    The deterministic model and dropout model both have a classification accuracy of 0\% in both the targeted and
    untargeted attack scenarios, indicated by the horizontal red dashed line at the bottom of the plot.
    This is the same accuracy on adversarial examples from this adversary reported in \citet{Carlini2016Adv} on a
    convolutional network trained on MNIST.
}
  \label{fig:adv_carlini_acc}
\end{figure*}

Figure~\ref{fig:adv_carlini_acc} plots the accuracy on $L_2$ optimization
adversarial examples of the first 1000 images from the MNIST
test set as a function of $\beta$.
The same sets of three models per $\beta$ were tested three times, as with the FGS
adversarial examples.

We generated both untargeted and targeted adversarial examples for
Figure~\ref{fig:adv_carlini_acc}.
For targeting, we generate a random target label different from the source label
in order to avoid biasing the results with unevenly explored source/target
pairs.
We see that for a reasonably broad range of $\beta$ values,
the VIB models have significantly better accuracy on the adversarial examples
than the deterministic models, which have an accuracy of 0\%
(the $L_2$ optimization attack is very effective on traditional model architectures).

Figure~\ref{fig:adv_carlini_acc} also reveals a surprising level of
adversarial robustness even when $\beta\rightarrow 0$.  This can be explained
by the theoretical framework of \citet{Fawzi2016}.  Their work proves
that quadratic classifiers (e.g., $x^{\tee}Ax$, symmetric $A$) have a greater
capacity for adversarial robustness than linear classifiers.
As we show in Appendix~\ref{sec:quad},
our Gaussian/softmax encoder/decoder is approximately quadratic
for all $\beta<\infty$.

\subsubsection{ImageNet Results and Discussion}
\label{sec:imagenet_experiments}

\begin{figure}[htbp!]
  \begin{center}
    \begin{tabular}{cc}
      \includegraphics[clip, trim=0px 222px 0px 0px, width=0.45\linewidth]{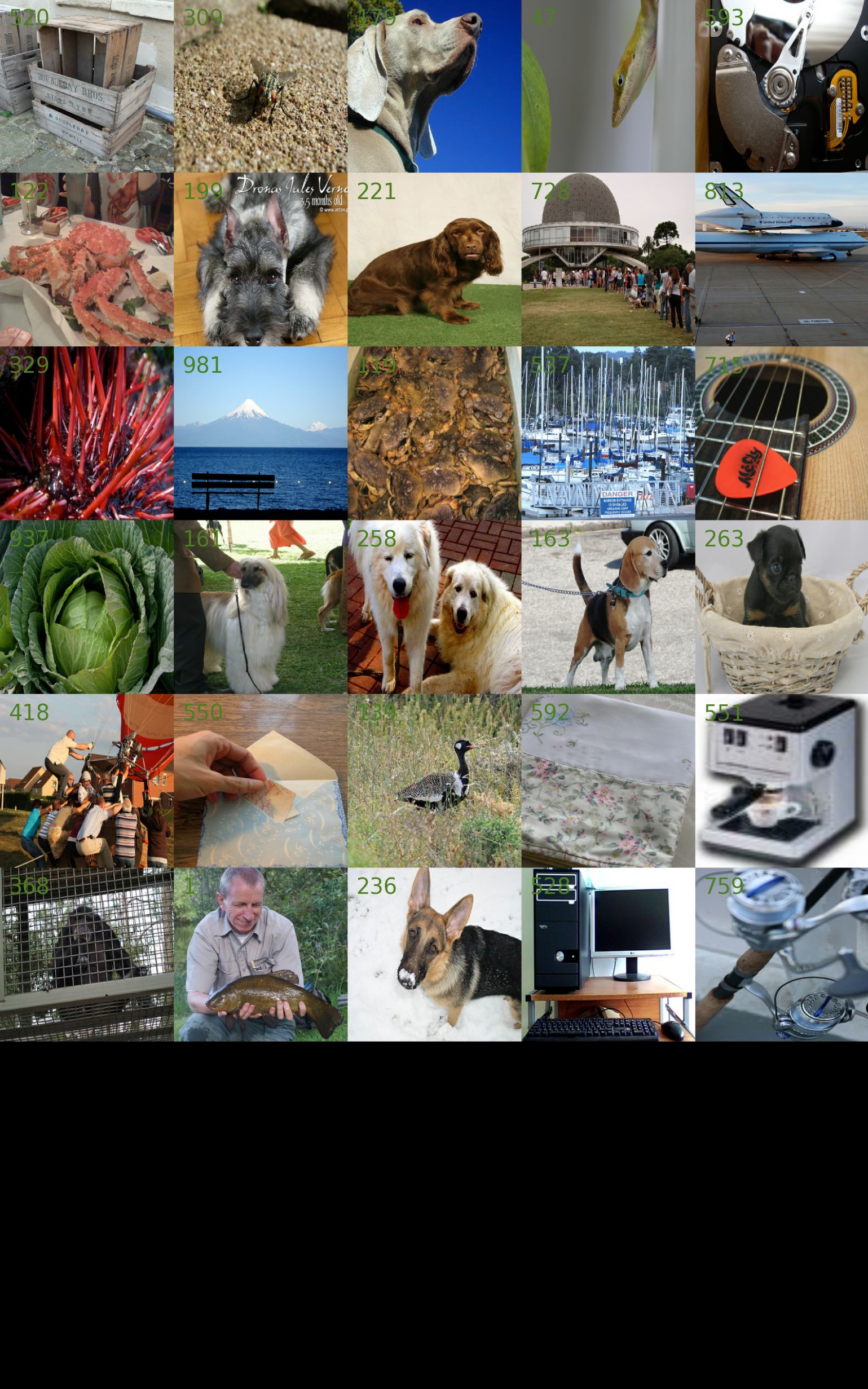}
      &
      \includegraphics[clip, trim=0px 222px 0px 0px,width=0.45\linewidth]{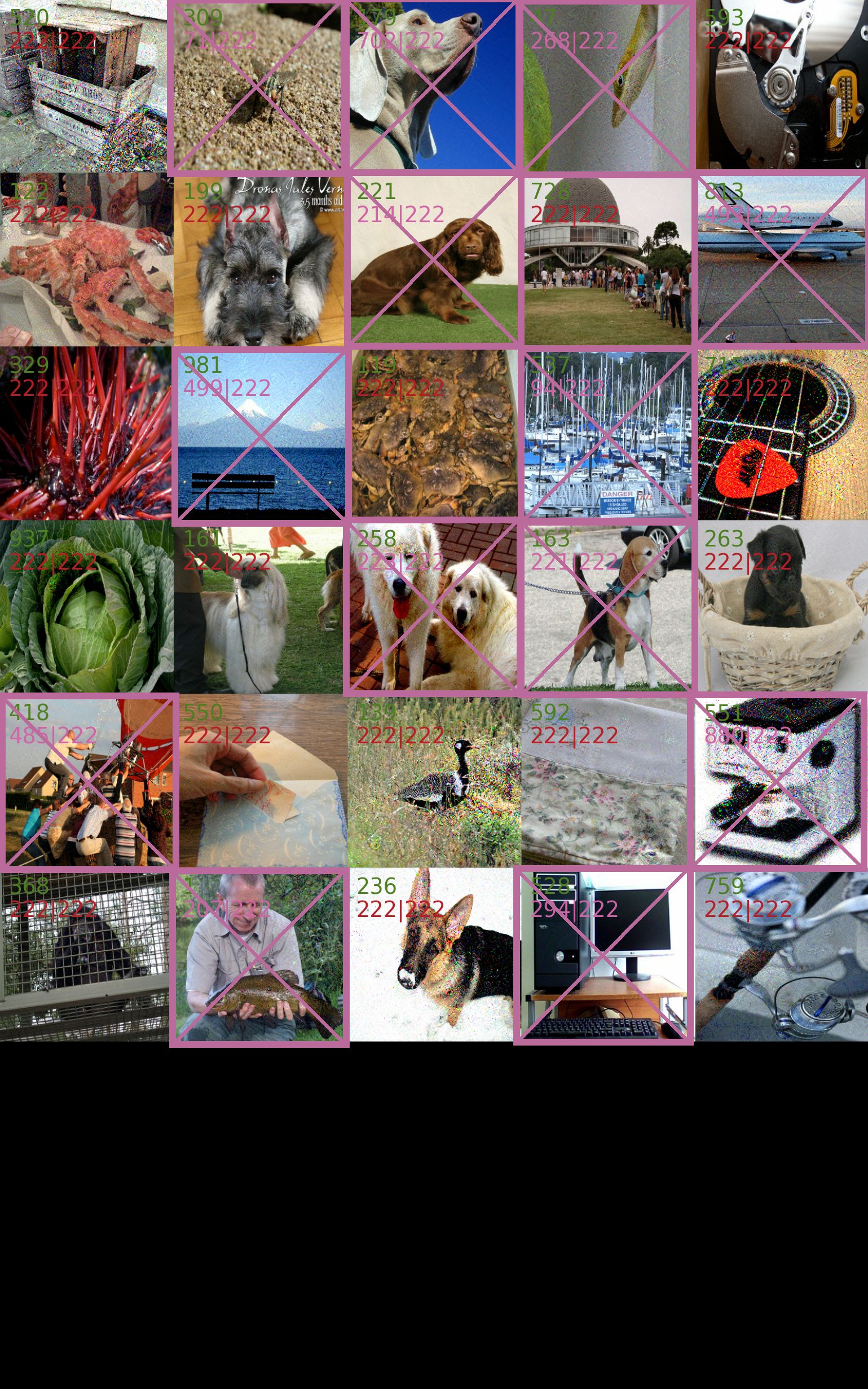} \\
      (a) & (b) \\
      \includegraphics[clip, trim=0px 222px 0px 0px,width=0.45\linewidth]{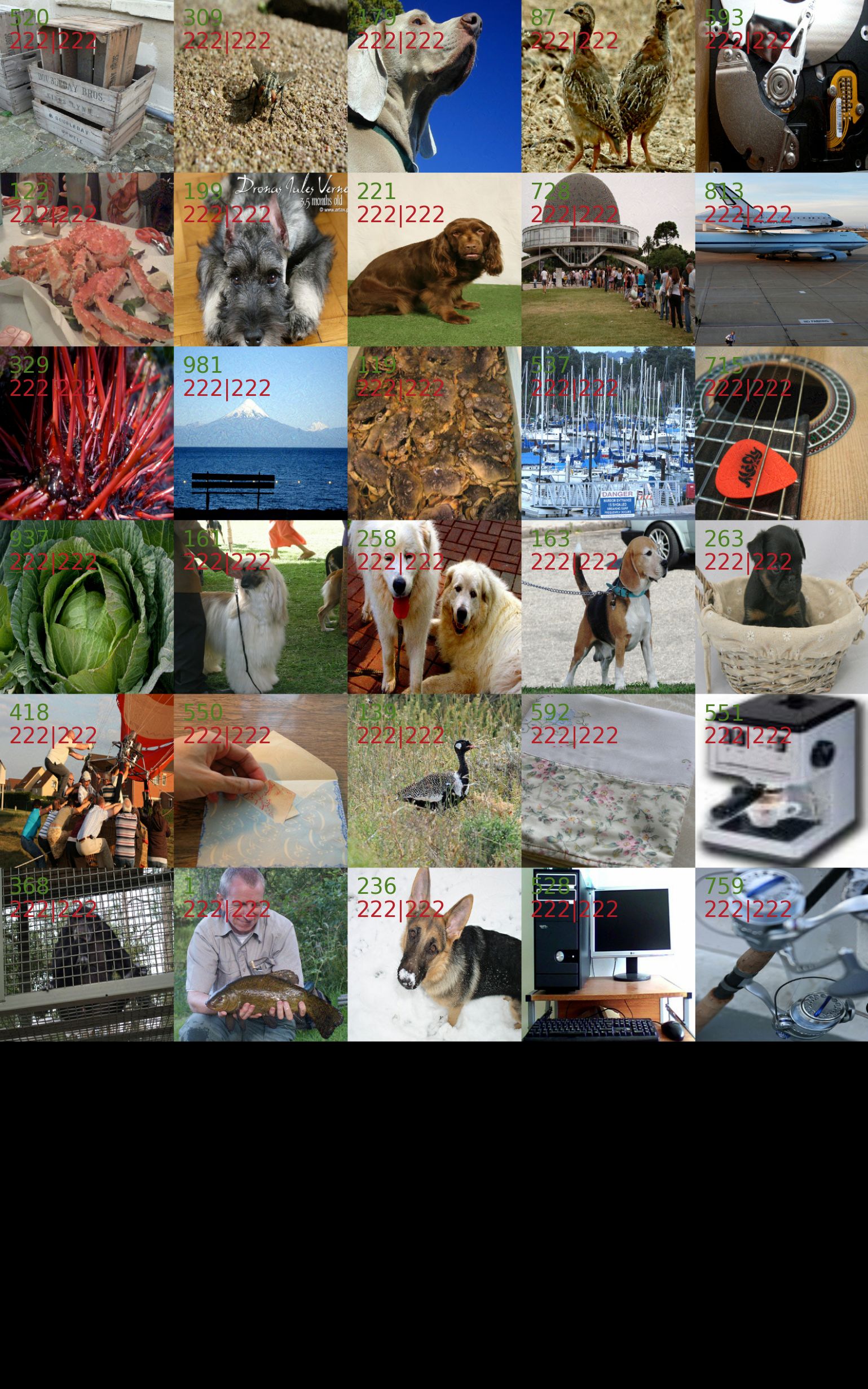}
    &
      \includegraphics[clip, trim=0px 222px 0px 0px,width=0.45\linewidth]{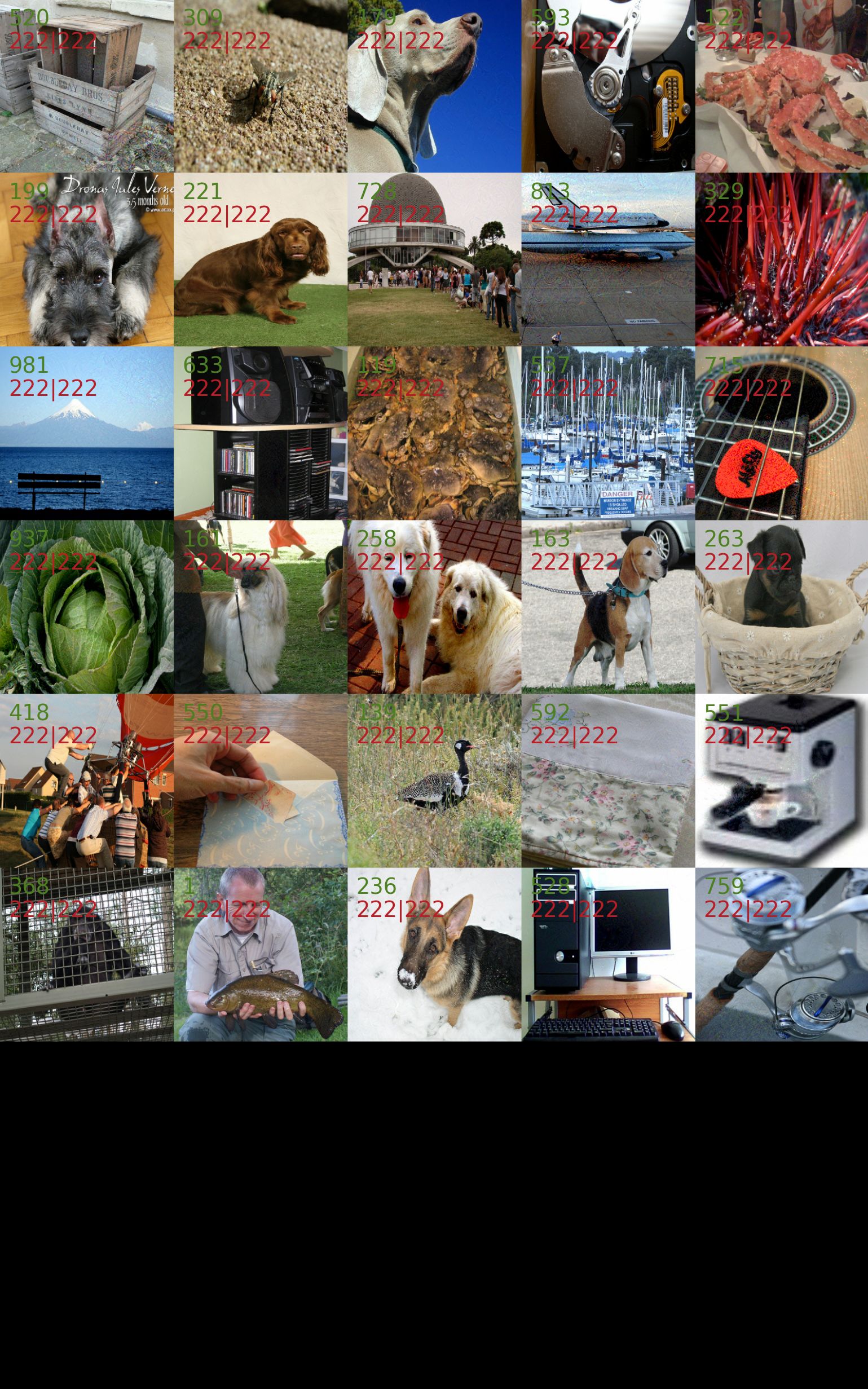} \\
        \\
        (c) & (d)
  \end{tabular}
  \caption{
    The results of our ImageNet targeted $L_2$ optimization attack.
    In all cases we target a new label of 222 (``soccer ball'').
    Figure (a) shows the 30 images from the first 40 images in the ImageNet validation set that the VIB network
    classifies correctly.
    The class label is shown in green on each image.
    The predicted label and targeted label are shown in red.
    Figure (b) shows adversarial examples of the same images generated by attacking our VIB network with $\beta=0.01$.
    While all of the attacks change the classification of the image, in 13 out of 30 examples the attack fails to hit
    the intended target class (``soccer ball'').
    Pink crosses denote cases where the attack failed to force the model
    to misclassify the image as a soccer ball.
    Figure (c) shows the same result but for our deterministic baseline operating on the whitened precomputed features.
    The attack always succceeds.
    Figure (d) is the same but for the original full Inception ResNet V2 network without modification.
    The attack always succceeds.
    There are slight variations in the set of adversarial examples shown for each network because we limited the
    adversarial search to correctly classified images. In the case of the deterministic baseline and original Inception ResNet V2 network,
	  the perturbations are hardly noticable in the perturbed images, but in many instances, the perturbations for the VIB network
	  can be perceived.
    }
	\label{fig:imagenetadv}
\end{center}
\end{figure}

\begin{figure}[htbp!]
  \begin{center}
  \includegraphics[clip, trim=0px 222px 0px 0px,width=0.95\linewidth]{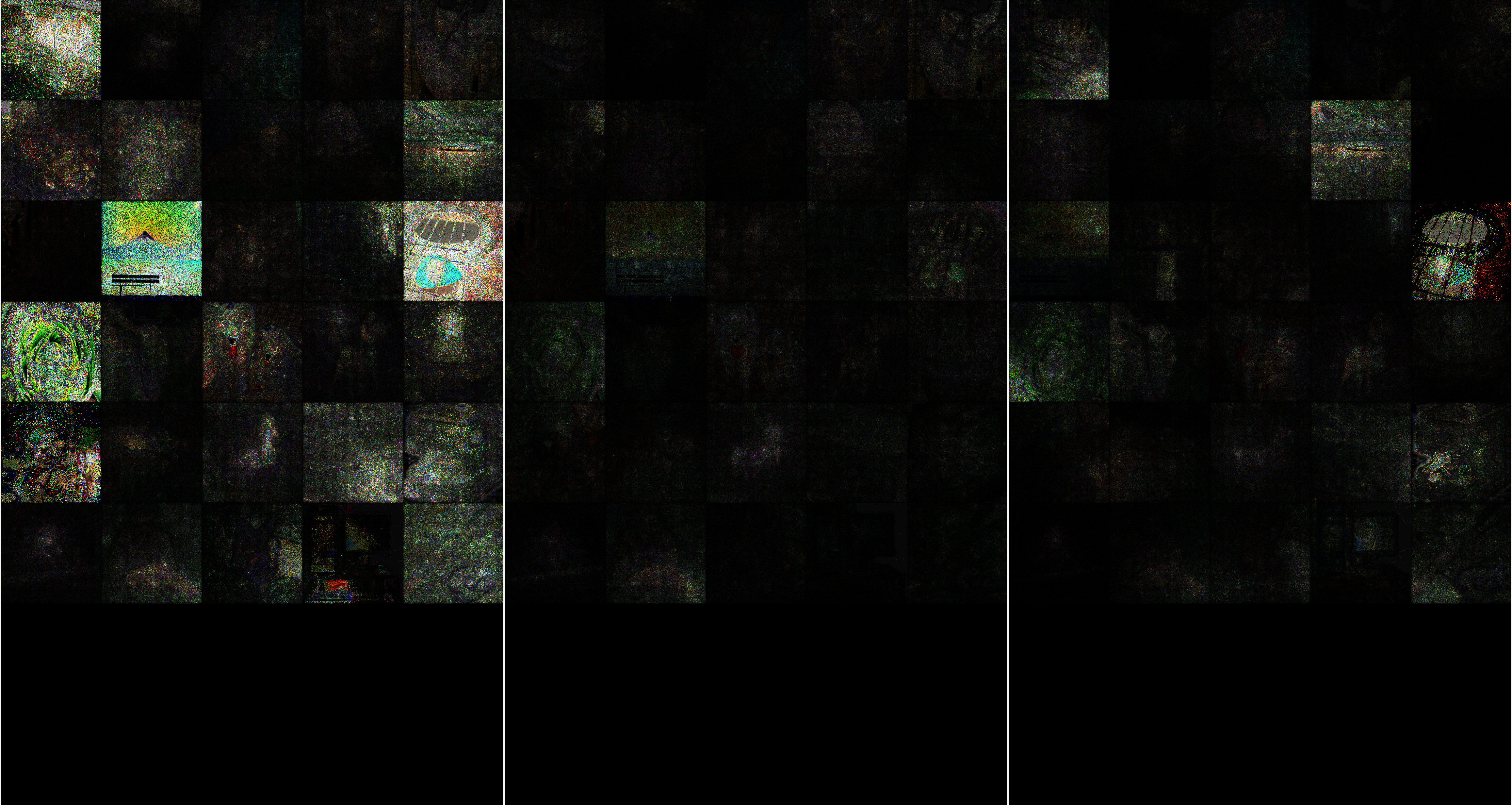}
  \caption{
    Shown are the absolute differences between the original and final perturbed images for all three networks.
    The left block shows the perturbations created while targeting the VIB network.
    The middle block shows the perturbations needed for the deterministic baseline using precomputed whitened
    features.
    The right block shows the perturbations created for the unmodified Inception ResNet V2 network.
    The contrast has been increased by the same amount in all three columns to emphasize the difference in the
    magnitude of the perturbations. The VIB network required much larger perturbations to confuse the classifier, and even then
	  did not achieve the targeted class in 13 of those cases.
  }
  \label{fig:imagenetcomp}
\end{center}
\end{figure}

VIB improved classification accuracy and adversarial robustness for toy
datasets like MNIST.  We now investigate if VIB offers similar advantages for
ImageNet, a more challenging natural image classification. Recall that ImageNet
has approximately 1M images spanning 1K classes. We preprocess images such that
they are 299x299 pixels.

\subsubsubsection{\it Architecture}

We make use of publicly available, pretrained
checkpoints\footnote{Available at the Tensorflow Models repository in the Slim
directory: \url{https://github.com/tensorflow/models/tree/master/slim}} of
Inception Resnet V2~\citep{inceptionresnet} on ImageNet~\citep{imagenetpaper}.
The checkpoint obtains 80.4\% classification accuracy on the ImageNet
validation set. Using the checkpoint, we transformed the original training set
by applying the pretrained network to each image and extracting the
representation at the penultimate layer. This new image representation has 1536
dimensions. The higher layers of the network continue to classify this
representation with 80.4\% accuracy; conditioned on this extraction the
classification model is simply logistic regression.  To further speed training,
we whitened the 1536 dimensional representation.

Under this transformation, the experiment regime is identical to the
permutation invariant MNIST task.  We therefore used a similar model
architecture.  Inputs are passed through two fully connected layers, each with
1024 units. Next, data is fed to a stochastic encoding layer; this layer is
characterized by a spherical Gaussian with 1024 learned means and standard
deviations. The output of the stochastic layer is fed to the variational
classifier--itself a logistic regression, for simplicity.  All other
hyperparameters and training choices are identical to those used in MNIST, more details in
Appendix~\ref{sec:hyperparameters}.

\subsubsubsection{\it Classification}

We see the same favorable VIB classification performance in ImageNet as in
MNIST\@.  By varying $\beta$, the estimated mutual information
between encoding and image ($I(Z,X)$) varies as well. At large values of
$\beta$ accuracy suffers, but at intermediate values we obtain improved
performance over both a deterministic baseline and a $\beta=0$ regime.  In all
cases our accuracy is somewhat lower than the original 80.4\% accuracy.
This may be a consequence of inadequate training time or suboptimal hyperparameters.

Overall the best accuracy we achieved was using $\beta=0.01$. Under this
setting we saw an accuracy of 80.12\%--nearly the same as the state-of-the-art unmodified network-- but with
substantially smaller information footprint, only $I(X,Z) \sim 45$ bits.  This
is a surprisingly small amount of information; $\beta=0$ implies over 10,000
bits yet only reaches an accuracy of 78.87\%.  The deterministic baseline,
which was the same network, but without the VIB loss and a 1024 fully connected
linear layer instead of the stochastic embedding similarly only achieved
78.75\% accuracy.  We stress that regressions from the achievable 80.4\% are
likely due to suboptimal hyperparameters settings or inadequate training.

Considering a continuum of $\beta$ and a deterministic baseline, the best
classification accuracy was achieved with a $\beta=0.01\in(0,1)$.  In other
words, VIB offered accuracy benefit yet using a mere $\sim 45$ bits of
information from each image.

\subsubsubsection{\it Adversarial Robustness}

We next show that the VIB-trained network improves resistance to adversarial
attack.

We focus on the Carlini targeted $L_2$ attack (see Section~\ref{sec:typesadv}).
We show results for the VIB-trained network and a deterministic baseline
(both on top of precomputed features), as well as for the original pretrained
Inception ResNet V2 network itself.
The VIB network is more robust to the targeted $L_2$ optimization attack in
both magnitude of perturbation and frequency of successful attack.

Figure~\ref{fig:imagenetadv} shows
some example images which were all
misclassified as ``soccer balls'' by the deterministic models;
by contrast, with the VIB model, only 17 out of 30 of the attacks succeeded
in being mislabeled as the target label.\footnote{
The attacks still often cause the VIB model to misclassify the image,
but not to the targeted label.
This is a form of ``partial'' robustness, in that an attacker will have
a harder time hitting the target class, but can still disrupt correct
function of the network.
}
We find that the VIB model can resist about
43.3\% of the attacks,
but the deterministic models always fail
(i.e., always misclassify into the targeted label).

Figure~\ref{fig:imagenetcomp} shows the absolute
pixel differences between the perturbed and unperturbed images
for the examples in Figure~\ref{fig:imagenetadv}.
We see that the VIB network
requires much larger perturbations in order to fool the classifier,
as quantified in Table~\ref{tab:imagenet}.

\begin{table}[htbp]
	\begin{center}
		\begin{tabular}{r|l|l|l}
			Metric 				& Determ & IRv2	 	& VIB(0.01)		 \\
			\hline
			Sucessful target 	& 1.0	 & 1.0 	 	& \textbf{0.567} \\
			$L_2$ 				& 6.45	 & 14.43 	& \textbf{43.27} \\
			$L_\infty$ 			& 0.18	 & 0.44 	& \textbf{0.92}  \\
		\end{tabular}
	  \caption{Quantitative results showing how the different Inception Resnet V2-based architectures
		(described in Section~\ref{sec:imagenet_experiments}) respond to targeted $L_2$ adversarial examples.
		\textit{Determ} is the deterministic architecture, \textit{IRv2} is the unmodified Inception
		Resnet V2 architecture, and \textit{VIB(0.01)} is the VIB architecture with $\beta=0.01$.
		\textit{Successful target} is the fraction of adversarial examples that caused the architecture
		to classify as the target class (soccer ball).  Lower is better.
		$L_2$ and $L_\infty$ are the average $L$ distances between the original images and the adversarial examples.
		Larger values mean the adversary had to make a larger perturbation to change the class.
	  }
	    \label{tab:imagenet}
	\end{center}
\end{table}

\section{Future Directions}
\label{sec:future}

There are many possible directions for future work, including:
putting the VIB objective at multiple or every layer of a network;
testing on real images;
using  richer parametric
marginal approximations, rather than assuming $r(z)=\gauss(0,I)$;
exploring the connections to differential privacy
(see e.g., \cite{Wang2016,Cuff2016});
and investigating open universe classification problems
(see e.g., \cite{Bendale2015}).
In addition, we would like to explore applications to sequence prediction,
where $X$ denotes the past of the sequence and $Y$ the future,
while $Z$ is the current representation of the network.  This form of the
information bottleneck is known as predictive information
\citep{bialek2001predictability,palmer2015predictive}.

\eat{
Potentially, implicit encodings could also be explored, though this makes bounding $I(Z,X)$ more difficult,
but by using variational approximations to the ratio of the marginal and conditional distributions, similar
to those used by Generative Adversarial Networks, the stochastic embeddings can potentially be made much richer.

Alternative variational bounds on the unsupervised objectives should be considered.

Additional, the Information Bottleneck is particularly attractive for
sequential data, where $X$ denotes the past of the sequence and $Y$ the future,
while $Z$ is the current representation of the network.  This form of the
information bottleneck is known as predictive information
\citep{bialek2001predictive}. Recently it has been shown that
populations of real sensory neurons seem to saturate this bound \citep{palmer2015predictive}.
}

\bibliography{bib}
\bibliographystyle{iclr2017_conference}

\newpage
\appendix
\section{Hyperparameters and Architecture Details for Experiments}
\label{sec:hyperparameters}

All of the networks for this paper were trained using
TensorFlow~\citep{tensorflow}.  All weights were initialized using the default
TensorFlow Xavier initialization scheme~\citep{xavier} using the averaging fan
scaling factor on uniform noise.  All biases were initialized to zero.  The
Adam optimizer~\citep{kingma2014adam} was used with initial learning rate of
$10^{-4}$, $(\beta_1 = 0.5, \beta_2 = 0.999)$ and exponential decay,
decaying the learning rate by a factor of 0.97 every 2 epochs.  The networks
were all trained for 200 epochs total.  For the MNIST experiments, a batch size
of 100 was used, and the full 60,000 training and validation set was used for
training, and the 10,000 test images for test results.  The input images were
scaled to have values between -1 and 1 before fed to the network.

All runs maintained an exponential weighted average of the parameters during the training run;
these averaged parameters were used at test time.
This is in the style of Polyak averaging~\citep{polyak}, with a decay constant of 0.999.
Our estimate of mutual informations were measured in bits.
For the VIB experiments in all sections, no other form of regularization was used.

For the 256 dimensional gaussian embeddings of Section~\ref{sec:mnist256}, a linear layer of size
512 was used to create the 256 mean values and standard deviations for the embedding. 
The standard deviations were made to be positive by a softplus transformation with 
a bias of -5.0 to have them initially be small.

\begin{equation}
	\sigma = \log\left(1 + \exp(x - 5.0)\right)
\end{equation}

For the 1024 dimensional Imagenet embeddings of Section~\ref{sec:imagenet_experiments}, a
sigma bias of 0.57 was used to keep the initial standard deviations near 1
originally, and a batch size of 200 was used.

For the 2 dimensional gaussian embeddings of Section~\ref{sec:mnist2}, a linear
layer was used with 2+4 = 6 outputs, the first two of which were used for the
means, and the other 4 were reshaped to a $2 \times 2$ matrix, the center was
transformed according to a softplus with a bias of -5.0, and the off diagonal
components were multiplied by $10^{-2}$, while the upper triangular element was
dropped to form the Cholesky decomposition of the covariance matrix.

\section{Connection to Variational Autoencoders}
\label{sec:vae}

We can also consider unsupervised versions of the information bottleneck objective.
Consider the objective:
\begin{equation}
	\max I(Z,X) - \beta I(Z, i),
\end{equation}
similar to the information theoretic objective for clustering introduced in \citet{unsupib}.

Here the aim is to take our data $X$ and maximize the mutual information contained
in some encoding $Z$, while restricting how much information we allow our
representation to contain about the identity of each data element
in our sample $(i)$.
We will form a bound much like we did in the main text.
For the first term, we form a variational decoder $q(x|z)$ and take a bound:

\begin{align}
	I(Z,X) &= \int dx\, dz\, p(x,z) \log \frac{p(x|z)}{p(x)} \\
	&= H(x) + \int dz\, p(z) \int dx\, p(x|z) \log p(x|z) \\
	&\geq \int dz\, p(z) \int dx\, p(x|z) \log q(x|z) \\
	&= \int dx\, p(x) \int dz\, p(z|x) \log q(x|z).
\end{align}

Here we have dropped the entropy in our data $H(X)$ because it is out of our control
and we have used the nonnegativity of the Kullbach-Leibler divergence to replace our 
intractable $p(x|z)$ with a variational decoder $q(x|z)$.

Turning our attention to the second term, note that:
\begin{equation}
	p(z|i) = \int dx\, p(z|x) p(x|i) = \int dx\, p(z|x) \delta(x- x_i) = p(z|x_i),
\end{equation}
and that we will take $p(i) = \frac 1 N$.

So that we can bound our second term from above
\begin{align}
	I(Z,i) &= \sum_i \int dz\, p(z|i) p(i) \log \frac{p(z|i)}{p(z)} \\
	&= \frac 1 N \sum_i \int dz\, p(z|x_i) \log \frac{p(z|x_i)}{p(z)} \\
	&\leq \frac 1 N \sum_i \int dz\, p(z|x_i) \log \frac{p(z|x_i)}{r(z)}, 
\end{align}
Where we have replaced the intractable marginal $p(z)$ with a variational marginal $r(z)$.

Putting these two bounds together we have that our unsupervised information bottleneck objective
takes the form
\begin{equation}
	I(Z,X) - \beta I(Z,i) \leq \int dx\, p(x) \int dz\, p(z|x) \log q(x|z) - \beta \frac 1 N \sum_i \KL[ p(Z|x_i) , r(Z) ].
\end{equation}
And this takes the form of a variational autoencoder \citep{Kingma2014}, except with the second KL divergence term
having an arbitrary weight $\beta$.

It is interesting that while this objective takes the same mathematical form as
that of a Variational Autoencoder, the interpretation of the objective is very
different.  In the VAE, the model starts life as a generative model with a
defined prior $p(z)$ and stochastic decoder $p(x|z)$ as part of the model, and the encoder
$q(z|x)$ is created to serve as a variational approximation to the true
posterior $p(z|x) = p(x|z)p(z)/p(x)$.  In the VIB approach, the model is
originally just the stochastic encoder $p(z|x)$, and the decoder $q(x|z)$
is the variational approximation to the true $p(x|z) = p(z|x)p(x)/p(z)$ and
$r(z)$ is the variational approximation to the marginal $p(z) = \int dx\, p(x)
p(z|x)$. This difference in interpretation makes natural suggestions for novel 
directions for improvement.

This precise setup, albeit with a different motivation was recently explored in
\citet{earlyvisual}, where they demonstrated that by changing the weight of the
variational autoencoders regularization term, there were able to achieve latent
representations that were more capable when it came to zero-shot learning and
understanding "objectness". In that work, they motivated their choice to change
the relative weightings of the terms in the objective by appealing to notions
in neuroscience.  Here we demonstrate that appealing to the information
bottleneck objective gives a principled motivation and could open the door
to better understanding the optimal choice of $\beta$ and more tools
for accessing the importance and tradeoff of both terms.

Beyond the connection to existing variational autoencoder techniques, we note
that the unsupervised information bottleneck objective suggests new directions
to explore, including targetting the exact marginal $p(z)$ in the regularization
term, as well as the opportunity to explore tighter bounds on the first $I(Z,X)$ 
term that may not require explicit variational reconstruction.

%\input{regularizers}
%\input{log-likelihood}
%\section{Quadratic Activation}
\section{Quadratic bounds for stochastic logistic regression decoder}
\label{sec:quad}

Consider the special case when the bottleneck $Z$ is a multivariate Normal,
i.e., $z|x\sim N(\mu_x, \Sigma_x)$ where $\Sigma_x$ is a $K \times K$ positive
definite matrix. The parameters $\mu_x,\Sigma_x$ can be constructed from a deep
neural network, e.g.,
\begin{align*}
  \mu_x &= \gamma_{1:K}(x) \\
  \chol(\Sigma_x) &= \diag(\log(1+\exp(\gamma_{K+1:2K})))+\subtril(\gamma_{2K+1:K(K+3)/2}),
\end{align*}
where $\gamma(x)\in\mathbb{R}^{K(K+3)/2}$ is the network output of input $x$.

Suppose that the prediction is a categorical distribution computed as $\sm(Wz)$
where $W$ is a $C \times K$ weight matrix and $\log\sm(x)=x-\lse(x)$ is the
log-soft-max function with $\lse(x) = \log\sum_{k=1}^K \exp(x_k)$ being the
log-sum-exp function.

This setup (which is identical to our experiments) induces a classifier which
is bounded by a quadratic function, which is interesting
because the theoretical framework
\citet{Fawzi2016} proves that quadratic classifiers have greater
capacity for adversarial robustness than linear functions.

We now derive an approximate bound using second order Taylor series expansion
(TSE). The bound can be made proper via \citet{browne15sharpquadratic}.
However, using the TSE is sufficient to sketch the derivation.

Jensen's inequality implies that the negative log-likelihood soft-max is upper
bounded by:
\begin{align*}
  -\log \E \left[ \sm(WZ) \middle| \mu_x, \Sigma_x \right] 
  &\le -\E \left[ \log \sm(WZ)  \middle| \mu_x, \Sigma_x\right] \\
  &= -W\mu_x + \E \left[\lse(WZ) \middle| \mu_x, \Sigma_x\right] \\
  &= -W\mu_x + \E \left[\lse(Z) \middle| W\mu_x, W\Sigma_x\right].
\end{align*}

The second order Taylor series expansion (TSE) of $\lse$ is given by,
\begin{align*}
  \lse(x+\delta) &\approx \lse(x) + \delta^\tee \sm(x) + \tfrac{1}{2} \delta^\tee \left[\diag(\sm(x)) - \sm(x)\sm(x)^\tee\right]\delta.
\end{align*}

Taking the expectation of the TSE at the mean yields,
\begin{align*}
\E_{N(0,W\Sigma_x W^\tee)}[&\lse(W\mu_x+\delta)] \approx \lse(W\mu_x) + \E_{N(0,W\Sigma_x W^\tee)}[\delta^\tee] \sm(W\mu_x) + \\
   &\hspace{2em}+\tfrac{1}{2} \E_{N(0,W\Sigma_x W^\tee)}[\delta^\tee \left[\diag(\sm(W\mu_x)) -
     \sm(W\mu_x)\sm(W\mu_x)^\tee\right]\delta] \\
&=\lse(W\mu_x) + \tfrac{1}{2} \tr(W\Sigma_x W^\tee \left[\diag(\sm(W\mu_x)) - \sm(W\mu_x)\sm(W\mu_x)^\tee\right]) \\
&=\lse(W\mu_x) + \tfrac{1}{2} \tr(W\Sigma_x W^\tee \diag(\sm(W\mu_x))) - \tfrac{1}{2} \sm(W\mu_x)^\tee W\Sigma_x W^\tee \sm(W\mu_x)\\
&=\lse(W\mu_x) + \tfrac{1}{2} \sqrt{\sm(W\mu_x)}^\tee W\Sigma_x W^\tee \sqrt{\sm(W\mu_x)} - \tfrac{1}{2} \sm(W\mu_x)^\tee W\Sigma_x W^\tee \sm(W\mu_x)\\
\end{align*}
The second-moment was calculated by noting,
\[\E[X^\tee B X]= \E \tr(XX^\tee B) = \tr(\E[XX^\tee]B)=\tr(\Sigma B).\]

Putting this altogether, we conclude,
\begin{align*}
  \E \left[ \sm(WZ) \middle| \mu_x, \Sigma_x \right] \gtrapprox
    \sm(W\mu_x)\exp\left(
  -\tfrac{1}{2} \sqrt{\sm(W\mu_x)}^\tee W\Sigma_x W^\tee \sqrt{\sm(W\mu_x)}
  +\tfrac{1}{2} \sm(W\mu_x)^\tee W\Sigma_x W^\tee \sm(W\mu_x)
\right).
\end{align*}

As indicated, rather than approximate the $\lse$ via TSE, we can make a sharp,
quadratic upper bound via \citet{browne15sharpquadratic}. However this merely
changes the $\sm(W\mu_x)$ scaling in the exponential; the result is still
log-quadratic.

\end{document}